\documentclass{article}

\usepackage{microtype}
\usepackage{graphicx}
\usepackage{subfigure}
\usepackage{booktabs} 

\usepackage{hyperref}
\usepackage{url}

\usepackage[accepted]{icml2025}

\usepackage{amsmath}
\usepackage{amssymb}
\usepackage{mathtools}
\usepackage{amsthm}
\usepackage{multirow}
\usepackage{mathabx}
\usepackage{makecell}
\usepackage{xcolor}
\usepackage{colortbl}
\usepackage{wrapfig}
\usepackage{array}

\newcommand\Tstrut{\rule{0pt}{2.3ex}}         

\newcommand{\ours}{HTH}

\usepackage[capitalize,noabbrev]{cleveref}

\theoremstyle{plain}

\theoremstyle{definition}

\theoremstyle{remark}

\usepackage[textsize=tiny]{todonotes}

\icmltitlerunning{Pushing the Boundaries of State Space Models for Image and Video Generation}

\begin{document}

\twocolumn[
\icmltitle{Pushing the Boundaries of State Space Models for Image and Video Generation}

\icmlsetsymbol{equal}{*}

\begin{icmlauthorlist}
\icmlauthor{Yicong Hong$^{1}$}{}
\icmlauthor{Long Mai$^{1}$}{}
\icmlauthor{Yuan Yao$^{1,2}$}{}
\icmlauthor{Feng Liu$^{1}$}{} \\ \vspace{5pt}
\icmlauthor{$^{1}$Adobe Research, $^{2}$University of Rochester}{} \\ \vspace{5pt}
\icmlauthor{\tt\small{Project Page:}\;\;\tt\small\url{https://yiconghong.me/HTH}}{}
\end{icmlauthorlist}



\icmlkeywords{Machine Learning, ICML}

\vskip 0.3in
]



\printAffiliationsAndNotice{}  

\begin{abstract}

While Transformers have become the dominant architecture for visual generation, linear attention models, such as the state-space models (SSM), are increasingly recognized for their efficiency in processing long visual sequences. However, the essential efficiency of these models comes from formulating a limited recurrent state, enforcing causality among tokens that are prone to inconsistent modeling of N-dimensional visual data, leaving questions on their capacity to generate long non-causal sequences. In this paper, we explore the boundary of SSM on image and video generation by building the largest-scale diffusion SSM-Transformer hybrid model to date (5B parameters) based on the sub-quadratic bi-directional Hydra and self-attention, and generate up to 2K images and 360p 8 seconds (16 FPS) videos. 
Our results demonstrate that the model can produce faithful results aligned with complex text prompts and temporal consistent videos with high dynamics, suggesting the great potential of using SSMs for visual generation tasks.
\end{abstract}

\section{Introduction}
\label{sec:intro}

\begin{figure*}[t!]
  \centering
  \includegraphics[width=0.86\textwidth]{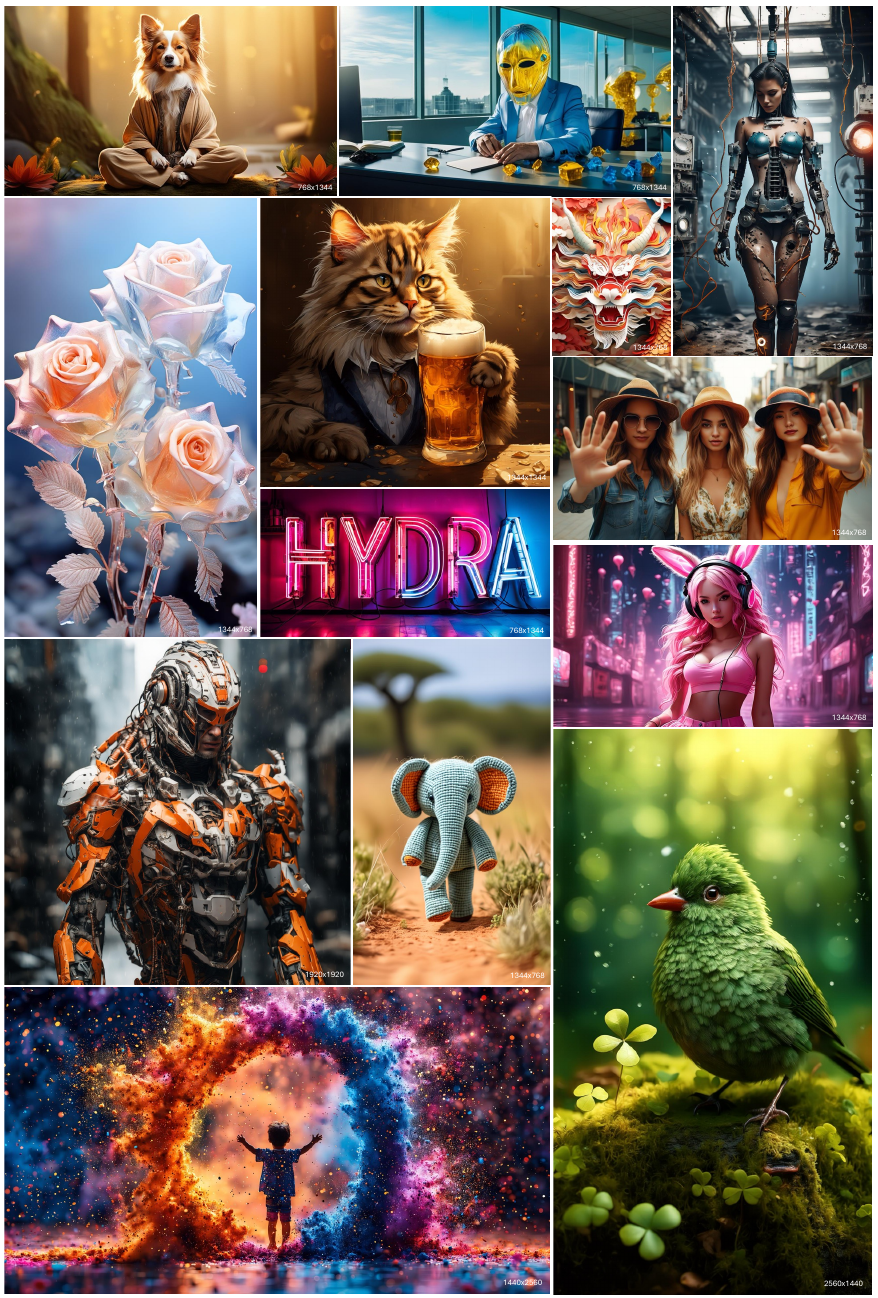}
  \vspace{-5pt}
  \caption{Text-to-1K/2K+ image generation results of our Hydra-Transformer Hybrid model. The resolution of each sample is displayed in the bottom-right corner. Text prompts and additional results are provided in the Appendix. Please zoom in for a clearer visualization.}
  \label{fig:visualize_1}
\end{figure*}

\begin{figure*}[t!]
  \centering
  \includegraphics[width=0.95\textwidth]{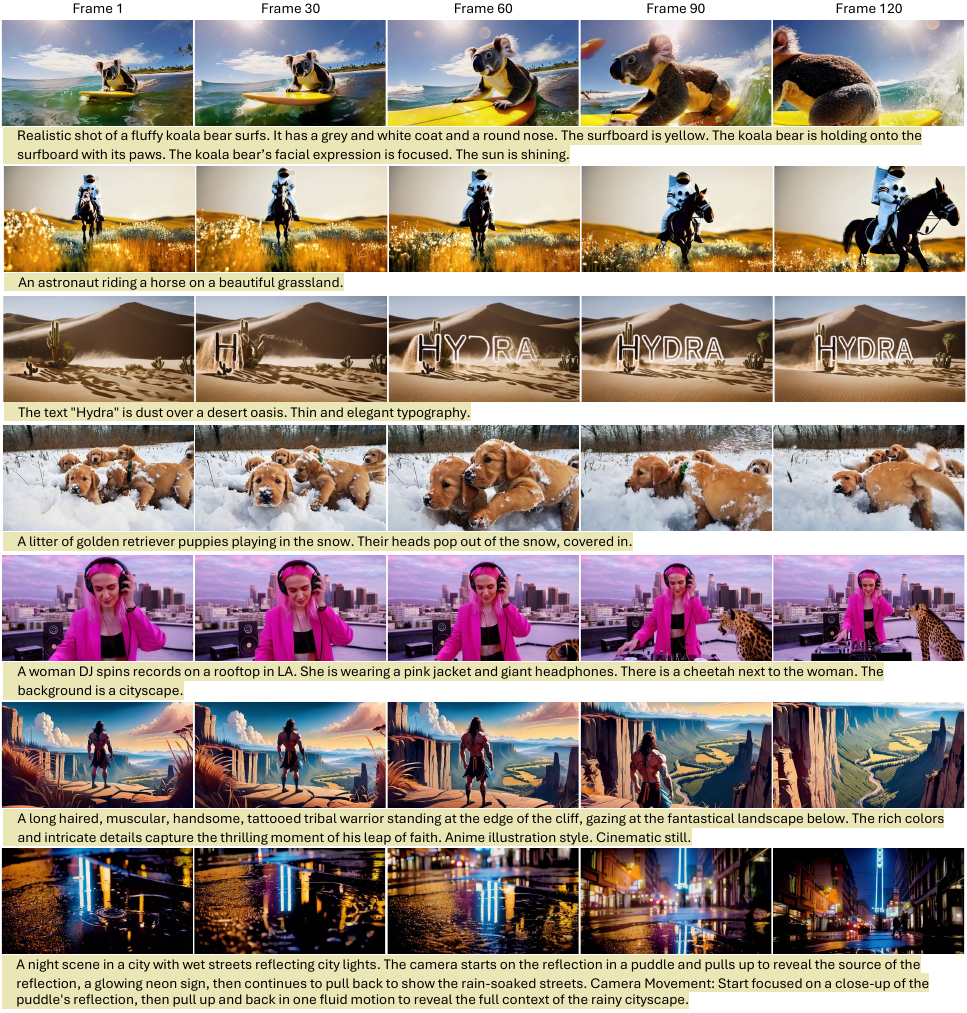}
  \vspace{-8pt}
  \caption{Text-to-360p 128 frames video generation results produced by our Hydra-Transformer Hybrid model. More results are provided in the Appendix. Please zoom in for a clearer visualization.}
  \label{fig:visualize_2}
  \vspace{-12pt}
\end{figure*}

In recent years, we have witnessed remarkable advancements in industrial-scale visual generation, spanning images~\citep{esser2021taming,dhariwal2021diffusion,rombach2022ldm,saharia2022imagen,peebles2023dit,chang2023muse,ruiz2023dreambooth,podell2023sdxl}, videos~\citep{yan2021videogpt,ho2022imagenvideo,blattmann2023align,girdhar2023emuvid,kondratyuk2023videopoet,bar2024lumiere}, 3D assets~\citep{hong2023lrm,li2023instant,xu2023dmv3d,wang2024dust3r,watson2024controlling}, embodied interactions~\citep{wu2023gr1,yang2023UniSim,du2024learning,cheang2024gr2,yang2024cogvideox}, and more. Central to many of these approaches is Transformer architecture~\citep{vaswani2017attention}, which serves as the core model backbone. Transformers, powered by the self-attention token mixing mechanism, effectively capture the relationships among visual tokens to produce realistic visual content and have proven highly scalable~\citep{dosovitskiy2020vit,dehghani2023scaling}. However, the quadratic computational complexity of Transformer models poses significant efficiency challenges, leading to dramatic slowdowns when processing long token sequences and making them impractical to common users~\citep{kitaev2020reformer,wang2020linformer,beltagy2020longformer}.

Meanwhile, there has been a surge in linear complexity models (LCM)~\citep{katharopoulos2020transformers,choromanski2020performers,shen2021efficient,qin2022cosformer,yang2023gated}; large models built based on RetNet~\citep{sun2023retentive}, RWKV~\citep{peng2023rwkv}, GLA~\citep{yang2023gated}, and Mamba~\citep{gu2023mamba,dao2024mamba2} have demonstrated performance comparable to Transformers across a variety of natural language processing tasks while achieving superior inference time speed. This success inspires recent research such as building linear complexity visual generative models~\citep{yan2023difussm,hu2024zigma,fei2024dimba,yang2023gated,zhu2024dig} and exploring Transformer-to-LCM distillation~\citep{wang2024mambainllama,bick2024mohawk,liu2024linfusion}. Despite their efficiency advantages, these methods often fail to produce visual outputs with the same quality as Transformer models. An essential reason behind this is that most of the LCMs are formulated as a stateful model that enforces causality on the input (tokens), which does not align with the non-causal N-dimensional nature of visual data. As a result, many recent generative visual LCMs implement multi-directional scans and preserve self-attention~\citep{gao2024matten,zhang2024motion,fei2024dimba,hu2024zigma,chen2024maskmamba,yi2024mvgamba}, which enlarges the receptive field of each token across layers.
However, those models and experiments mostly only deal with short visual sequences (\textit{e.g.}, low-res images), which cannot reflect the memory limitation or justify the expressiveness of LCMs~\citep{jelassi2024repeat,merrill2024illusion}. On the other hand, several works attempt to build non-casual LCMs by squeezing all visual tokens into the fix-sized state at once~\citep{liu2024linfusion,xie2024sana} or establishing a state representation that shares the same information with all the tokens~\citep{shi2024vssd}; despite encouraging results on image generation, our preliminary experiments show that these methods struggle with complex text prompts or video generation, where the latter requires learning much more complicated spatiotemporal patterns.

In light of the above, this paper takes an empirical approach to explore the boundary of State Space Models (SSM) in long-sequence visual generation with diffusion. Specifically, we apply the best-performing bi-directional SSM, Hydra, and follow the previous research to build a Hydra-and-Transformer Hybrid model (\textbf{\ours{}}). We consider the simplest horizontal and vertical raster scan patterns while focusing on scaling the model to 5B parameters for generating up to 2K images (1440$\times$2560) and 360p 8-second videos (16 FPS), respectively. Compared to 2D images, video data has an extra dimension which introduces large scanning complexity to the SSM. Here, to adapt the image-trained \ours{} base model for video generation, we propose a simple but effective method that directly changes the scanning order of some layers to temporal-first, forcing these layers to learn the temporal evolution of tokens at the same 2D position. 
Our final results reveal that the \ours{} model effectively generates faithful outputs that align with complex text prompts while producing temporally consistent and highly dynamic videos, highlighting the potential of sequential linear complexity models in visual generation.

\section{Background}
\label{sec:preliminaries}

\subsection{Selective State Space Models}
The recently proposed selective state space models (S4, \textit{e.g.}, Mamba~\citep{gu2023mamba,dao2024mamba2}) formulate a sequential transformation that maps a sequence $x_{t} \in \mathbb{R}^{d} \mapsto y_{t} \in \mathbb{R}^{d}$ through an implicit latent state $h_{t} \in \mathbb{R}^{N}$. Specifically, S4 models are defined with four input-dependent parameters, $\Delta_{t}$, $\mathbf{A}_{t}$, $\mathbf{B}_{t}$, and $\mathbf{C}_{t}$, which their discrete form can be generally expressed as 
\begin{equation}
\begin{aligned}
    h_{t} &= \overline{\mathbf{A}}_{t}h_{t-1} + \overline{\mathbf{B}}_{t}x_{t} \quad & 
    y_{t} &= \mathbf{C}^{\top}_{t}h_{t}
\end{aligned}
\end{equation}
where $\Delta_{t}$ controls the preservation of the current state, and $\overline{\mathbf{A}}_{t}$ and $\overline{\mathbf{B}}_{t}$ are produced by a discretization rule such as via the Zero-Order Hold as
\begin{equation}
\begin{aligned}
    \overline{\mathbf{A}}_{t} &= \text{exp}(\Delta_{t}\mathbf{A}_{t}) \\
    \overline{\mathbf{B}}_{t} &= (\Delta_{t}\mathbf{A}_{t})^{-1}(\text{exp}(\Delta_{t}\mathbf{A}_{t})-\mathbf{I})\cdot\Delta_{t}\mathbf{B}_{t}
\end{aligned}
\end{equation}
Referring to the state-space duality~\citep{dao2024mamba2,hwang2024hydra}, SSMs present a matrix transformation form for a $({\overline{\mathbf{A}},\overline{\mathbf{B}},\mathbf{C}})$-parameterized semiseparable matrix $\mathbf{M}_{\theta} \in \mathbb{R}^{(T,T)}$, where $T$ indicates the length of input sequence $\mathbf{X} \in \mathbb{R}^{(T,d)}$. Concisely, we can view the sequential transition
\begin{equation}
\begin{aligned}
    y_{t} &= \Sigma^{t}_{s=0}\mathbf{C}^{\top}_{t}\overline{\mathbf{A}}^{\times}_{t:s}\overline{\mathbf{B}}_{s}x_{s} \\
    \overline{\mathbf{A}}^{\times}_{i:j}&=\begin{cases}
    \prod^{i}_{k=j+1}\overline{\mathbf{A}}_{k},     & i>j \\
    1, & i=j
\end{cases}
\end{aligned}
\end{equation}
as
\begin{equation}
    \mathbf{Y}=\mathbf{M}_{\theta}\mathbf{X}
\end{equation}
where $m_{ij}=\mathbf{c}^{\top}_{i}\overline{\mathbf{A}}_{i}\cdot\cdot\cdot\overline{\mathbf{A}}_{j+1}\overline{\mathbf{b}}_{j}$ are the $(i,j)$-entries that fill the lower-triangular part of $\mathbf{M}_{\theta}$.

\subsection{Hydra: Bidirectional SSM}
A core limitation of the above SSM formulation lies in its inherent causality, which constrains its ability to model N-dimensional visual data effectively. To address this issue while allowing the model to possess strong expressiveness and leverage sub-quadratic matrix multiplication algorithms, the Hydra bidirectional SSM is introduced~\citep{hwang2024hydra}. Hydra employs a quasiseparable matrix as the token mixer and is implemented using the Mamba-2 framework~\citep{dao2024mamba2}.

Specifically, the quasiseparable matrix (\textit{QS}) in Hydra is formed by combining two semiseparable matrices (\textit{SS}) corresponding to forward and reverse sequence modeling:
\begin{equation}
    QS(\mathbf{X})=\hat{s}(SS(\mathbf{X}))+\hat{f}(\hat{s}(SS(\hat{f}(\mathbf{X}))))+\mathbf{D}\mathbf{X}
\end{equation}
where $\textbf{D}$ denotes the learnable diagonal parameters of the quasiseparable matrix. $\hat{f}$ and $\hat{s}$ are two non-parameterized operations; $\hat{f}$ reverses the input sequence, $\hat{s}$ shifts the sequence right by one position and pads the beginning with zero. As claimed and empirically justified in the Hydra paper, the above formulation of a quasiseparable matrix is strictly more expressive than existing addition-based bidirectional SSMs. We refer readers to the original paper for proof of the derivation~\citep{hwang2024hydra}.

\section{Method}
\label{sec:method}

\begin{figure*}[t]
  \centering
  \includegraphics[width=0.93\textwidth]{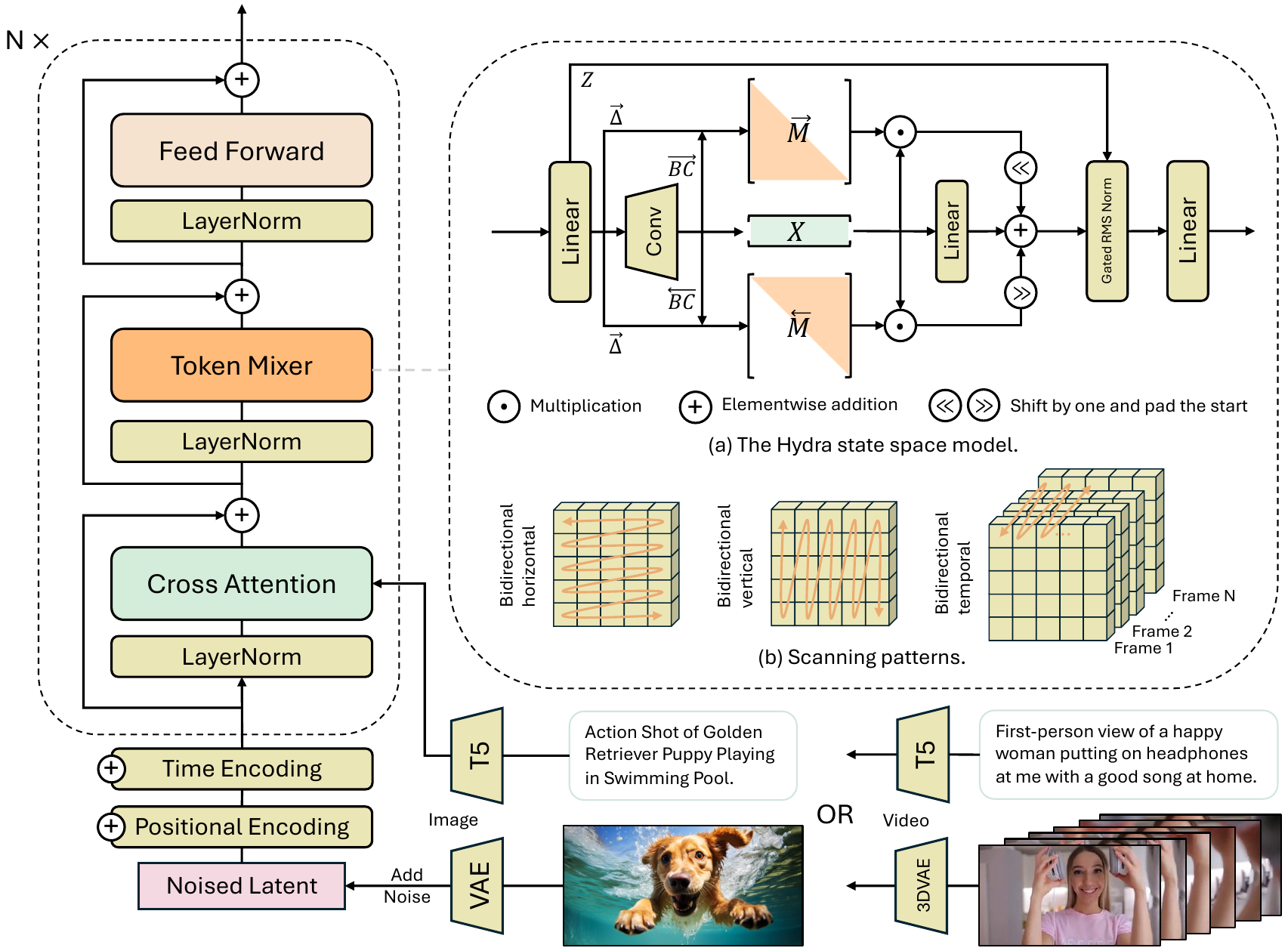}
  \vspace{-5pt}
  \caption{Illustration of our diffusion Hybrid Hydra (\ours{}) model for image and video generation. The architecture consists of $N$ stacked blocks, each comprising a cross-attention layer, a token mixer, and a feed-forward network. (a) The token mixer can be implemented as either the Hydra state space model or self-attention. (b) For image data, we use horizontal and vertical bidirectional raster scans on tokens, and for video data, an additional bidirectional temporal scan is applied.}
  \label{fig:hth_model}
  \vspace{-7pt}
\end{figure*}

We present our Hydra-Transformer Hybrid model (\ours{}), a diffusion framework for text-guided image and video generation (Figure~\ref{fig:hth_model}). Similar to the Diffusion Transformer~\citep{peebles2023dit}, we consider the simplest network components and architecture to establish a general and highly scalable large model. Inspired by recent study~\citep{waleffe2024empirical}, we effectively combine the Hydra state space model and self-attention by simply switching the token mixer across layers to allow our \ours{} to benefit from both efficient processing and global interaction. We specify the network architecture of \ours{} in \S\ref{subsec:hth} and \S\ref{subsec:integration}, and the adaptation of T2I to T2V training in \S\ref{subsec:adapt}.

\subsection{\ours{} Architecture}
\label{subsec:hth}

\paragraph{Data Encoders} 
The framework starts with textual and visual encoders for preprocessing the text prompt and its corresponding image or video. Similar to existing approaches~\citep{ho2022imagenvideo,kondratyuk2023videopoet,sun2024autoregressive,FLUX2024}, we adopt the T5 language model~\citep{raffel2020exploringt5}, specifically FLAN-T5-XXL~\citep{chung2024scalingt5}, to acquire textual representations. We compress raw visual input into a low-dimensional continuous latent space using a 3D VAE adapted from MAGVIT-v2~\citep{yu2023magvitv2}. The VAE can encode 2D images and incorporate motion encoding from video frames. Both VAEs produce 12-channel latent representations with an 8$\times$ spatial compression and $\frac{16}{5}\times$ temporal compression.

\paragraph{Denoising Model}
As visualized in the top-left of Figure~\ref{fig:hth_model}, the denoising model in \ours{} is formed by stacking $N$ blocks of residual-connected and pre-normed~\citep{xiong2020onlayer,lei2016layernorm} cross-attention, token mixer, and feed-forward layers. Starting with the cross-attention layer, it attends the input token sequence (\textit{i.e.}, the noised image/video latents) to the textual representations to acquire language guidance. Then, the token mixer layer, either Hydra or self-attention, models and coordinates the interactions among the visual latents, followed by a feed-forward layer to incorporate general knowledge. We choose Hydra among the SSM variants due to its efficient implementation of a non-causal model with superior expressiveness~\citep{hwang2024hydra}. We refer to Appendix~\ref{appsec:token_mixers} for comparison between unidirectional SSM, bidirectional SSM, the state-sharing VSSD~\citep{shi2024vssd}, self-attention, and Hydra.

\begin{figure*}[t]
  \centering
  \includegraphics[width=0.80\textwidth]{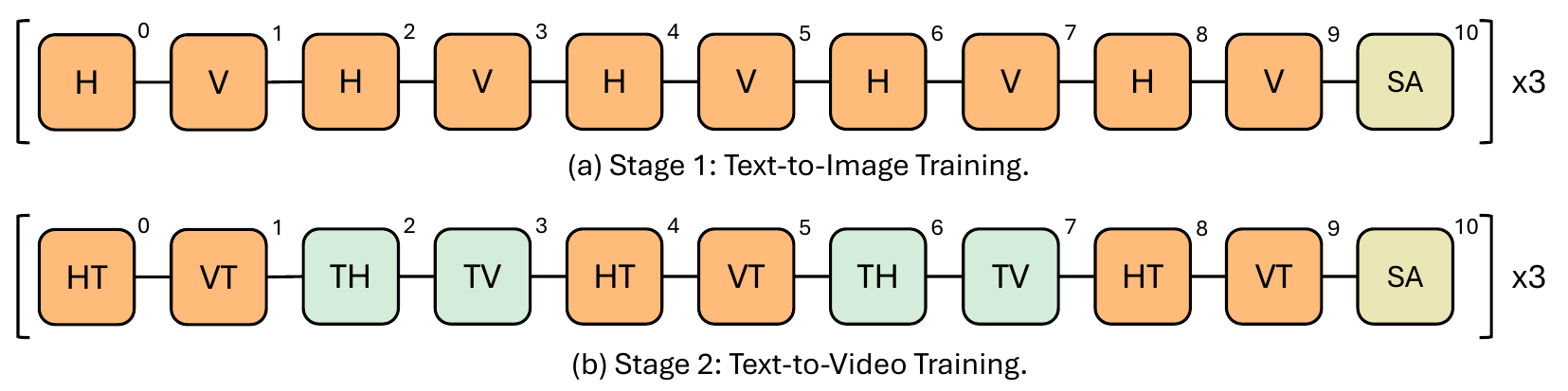}
  \vspace{-7pt}
  \caption{Illustration of the model adaptation from state 1 (T2I) to state 2 (T2V). For each set of 11 blocks in our \ours{} model, we change the scanning pattern of certain Hydra layers from spatial-major to temporal-major scan when processing video data.}
  \label{fig:hybrid_formula}
  \vspace{-10pt}
\end{figure*}

\paragraph{Scanning Patterns} Due to the bi-directional sequential nature of Hydra, the processed tokens are biased towards a higher correspondence to the adjacent tokens in the flattened 1D sequence, which contradicts the structure continuity of 2D and 3D visual data. To enlarge the receptive field of tokens, we alternate the scanning pattern between each of the two consecutive Hydra blocks. As shown in Figure~\ref{fig:hth_model}, we only consider two bi-directional raster scans, 
via either a horizontal or a vertical path. While previous research has explored other scanning patterns in various domains~\citep{zhang2024surveymamba,xu2024surveymamba}, their impact on large-scale diffusion SSMs remains unclear. Experimenting with different scanning patterns falls outside the scope of this paper and is left for future work.

\paragraph{Conditioning Mechanisms}
We leverage simple but effective methods to incorporate the diffusion and textual conditions. For the learnable diffusion timestep encoding, we add it directly to the noised latents once, before passing them to the first \ours{} layer. In contrast to the Adaptive LayerNorm applied in DiT~\citep{peebles2023dit}, this approach significantly saves the parameters and computation of the model. For the text prompts, we use a cross-attention layer to introduce semantic guidance from the text tokens to the visual tokens. Exploring other possible design choices, such as formulating in-context multimodal SSM (\textit{i.e.}, prepending text tokens to visual tokens, and initiating scanning from the text tokens~\citep{hu2024zigma}), is out of the scope of this paper. We leave this exploration for future work.

\subsection{Hydra and Self-Attention Integration}
\label{subsec:integration}
Inspired by existing works that combine SSM layers and Transformer layers~\citep{zuo2022augmented,waleffe2024empirical,lieber2024jamba,fei2024dimba,ziwen2024longlrm,wang2024mambainllama} as well as other linear hybrid RNN networks~\citep{botev2024recurrentgemma,de2024griffin}, we create our hybrid model that enjoys the efficiency advantage and global interaction from SSM and self-attention, respectively. As shown in Figure~\ref{fig:hybrid_formula}(a), our \ours{} model consists of a total $N{=}33$ blocks, with each set of 11 blocks containing 10 Hydra blocks followed by a final self-attention block. We formulate \ours{} to have a high Hydra-to-Transformer ratio ($30{:}3$) to exploit the modeling capacity of SSMs and preserve efficiency.

\subsection{T2I to T2V Adaptation}
\label{subsec:adapt}
Following the common multi-stage training strategy~\citep{hong2022cogvideo,blattmann2023align,girdhar2023emuvid,bar2024lumiere}, we first train a base T2I model before applying video data, allowing the T2V \ours{} to inherit the rich semantics and high visual quality learned from abundant images. However, video data introduces an extra dimension that significantly increases the modeling distance between spatially or temporally adjacent tokens and demands additional temporal-scanning SSMs to model the frame-wise consistency and evolution. Our early empirical results indicate that neither directly using the stage 1 spatial-major scanning (Figure~\ref{fig:hybrid_formula}(a)) nor adding new temporal-major scanning Hydra blocks is effective for video data; we hypothesize that the former results in a weak temporal receptive field for the tokens, while the latter introduces excessive noise, hindering learning. 

Through extensive experiments, we found a simple but surprisingly effective method -- revising the scanning patterns of certain spatial-major scanning blocks to temporal-major scanning. As illustrated in Figure~\ref{fig:hth_model}(b) and Figure~\ref{fig:hybrid_formula}(b), we directly change every two Hydra blocks with horizontal and vertical scans ($\texttt{H}$ and $\texttt{V}$) to temporal-horizontal and temporal-vertical scans ($\texttt{TH}$ and $\texttt{TV}$, \textit{i.e.}, the SSMs scan all tokens at the same temporal position before moving to a horizontal or vertical adjacent token at the first frame), while having the other blocks to perform spatial-temporal frame-by-frame scanning ($\texttt{HT}$ and $\texttt{VT}$). Note that for image data, temporal scans are absent, so all layers function as in stage 1. Therefore, this approach actually forces 40\% of the Hydra blocks to learn additional temporal-major modeling.

\section{Experiments}
\label{sec:experiments}

We provide the implementation details of our model, along with the results of text-to-image and text-to-video generation at different resolutions.

\paragraph{Implementation Details} 
As mentioned in \S\ref{subsec:integration}, our \ours{} model comprises 33 blocks, with every 10 Hydra blocks followed by a self-attention block. The model uses a hidden dimension of 3072 throughout, with Hydra configured to have an internal state dimension of 256, \texttt{ngroups} set to 1, a head dimension of 64, an expansion factor of 2$\times$, and a convolution window size of 7. For both cross-attention and self-attention layers, the KV-dimension matches the model dimension (3072), utilizing 32 attention heads and incorporating RMSNorm~\citep{zhang2019rmsnorm} for QK-Normalization~\citep{henry2020qknorm}. The MLP in each block applies an expansion ratio of 2$\times$ and consists of two nonlinear layers. Together, the above configurations result in the final 5B-parameter \ours{} model.
As illustrated in Figure~\ref{fig:hth_model}, residual connections are employed at the layer level, linking features preceding each LayerNorm~\citep{lei2016layernorm} to the output features of the subsequent layer.
We use a patch size of 2$\times$2 to further compress the spatial dimension of latents produced by the VAE. To incorporate positional information, we use a non-learnable absolute sinusoidal positional embedding~\citep{vaswani2017attention}, which is directly added to the input tokens.

\paragraph{Training} 
We adopt a multi-stage training strategy to enable high-resolution image and video generation. Initially, we train a base T2I model on 256p internal data for 200K iterations with a batch size of 1024, followed by an additional 286K iterations with an increased batch size of 3072. The starting learning rate is set to 1e-4 and decays according to a cosine annealing schedule. Next,
\begin{itemize}
    \item \textbf{1K T2I}. We apply a fast adaptation training to extend the model for high-resolution image generation. Specifically, we fine-tune the 256p base model on 1K-resolution images using a batch size of 576, training for only 10,000 iterations with a learning rate that rapidly decays from 5e-5 to 1e-6.
    \item \textbf{180p T2V}. To obtain a base video generation model, we continue training the T2I base model on a mixture of 256p images and 180p videos (characterized by a smaller number of frames on average and varying FPS) over 154K iterations. 
    We adopt a stepwise learning rate reset approach, where the starting learning rate of the cosine annealing schedule is reset whenever the validation loss plateaus. In our experiment, the learning rate is reset four times: from 1e-4 to 5e-5, then to 3e-5, and finally to 1e-5.
    This training stage enables the model to synthesize video data while preserving the learned semantics and visual quality. 
    \item \textbf{360p T2V}. Finally, we tune the 180p video model using 360p 128-frame video data with a batch size of 288. Leveraging a similar fast adaptation training strategy, we train the model for only 10K iterations with a learning rate that decays from 1e-5 to 1e-6.
\end{itemize} 
We apply the diffusion noise scheduling and loss introduced in InstaFlow~\citep{liu2023instaflow} and train the model following standard diffusion model training procedure~\citep{ho2020ddpm} with classifier-free guidance~\citep{ho2022classifier}.

\paragraph{Inference} 
At inference, we apply DDIM~\citep{song2020ddim} to denoise the latent representations. The denoised latents are then decoded by our internal VAE, with the resulting outputs presented and evaluated in this paper without any post-processing, such as super-resolution.

\subsection{Results and Discussions}

\subsubsection{Visualization} We present several 1K$+$ images and 360p video frames generated by our \ours{} model in Figure~\ref{fig:visualize_1} and Figure~\ref{fig:visualize_2}, respectively. Additional results, including playable videos, are provided in Appendix~\ref{appsec:visualization} and the supplemental materials. The visualizations demonstrate that the model can produce high-fidelity outputs across diverse styles consistent with the given text prompts. Notably, for high-res image generation, despite being trained on mixed-aspect ratios 1K data (with extreme \texttt{Height:Width} ratios of 512:2048 and 2048:512), the model is capable of zero-shot generating around 3.5$\times$ larger 2K images, including resolutions up to 1920$\times$1920, 1440$\times$2560, and 2560$\times$1440. We refer readers to Appendix~\ref{appsec:highres_zeroshot} for more details about the zero-shot experiments.

\begin{table}[t]
  \caption{Comparison of 256$\times$256 T2I generation.}
  \vspace{3pt}
  \begin{center}
  \resizebox{0.67\columnwidth}{!}{
    \begin{tabular}{lc}
        \hline \hline & \\[-2.0ex]
        \multicolumn{1}{c}{\multirow{2}{*}{Models}} & \multicolumn{1}{c}{MSCOCO-30K} \\
        \cline{2-2} & 
        \multicolumn{1}{c}{FID$\downarrow$}\Tstrut\\
        \hline \hline & \\[-2.0ex]
        LDM~\citep{rombach2022ldm}              & 12.63 \\
        DALL-E 2~\citep{ramesh2022dalle2}       & 10.39 \\
        Imagen~\citep{saharia2022imagen}        & 7.27  \\
        eDiff-I~\citep{balaji2022ediff}         & 6.95  \\
        Transfusion~\citep{zhou2024transfusion} & 6.78  \\
        DeepFloyd~\citep{deepfloyd2024}         & 6.66  \\
        RAPHAEL~\citep{xue2024raphael}          & 6.61  \\
        \hline & \\[-2.0ex]
        DiT (Ours, baseline)  & 12.53 \\
        HTH (Ours)  & 11.85 \\
        \hline \hline
      \end{tabular}
        }
  \end{center}
  \label{tab:eval_256p}
  \vspace{-15pt}
\end{table}

\begin{table}[t]
  \caption{Comparison of 1024$\times$1024 T2I generation.}
  \vspace{-5pt}
  \begin{center}
  \resizebox{\columnwidth}{!}{
    \begin{tabular}{lccc}
        \hline \hline & \\[-2.0ex]
        \multicolumn{1}{c}{\multirow{2}{*}{Models}} & \multicolumn{2}{c}{MJHQ-30K} & \multicolumn{1}{c}{GenEval} \\
        \cline{2-4} & 
        \multicolumn{1}{c}{FID$\downarrow$} & \multicolumn{1}{c}{CLIP-Score$\uparrow$} &
        \multicolumn{1}{c}{Overall$\uparrow$} \Tstrut\\
        \hline \hline & \\[-2.0ex]
        SDXL~\citep{podell2023sdxl}      &  8.76 & 28.60  & 0.55 \\
        PixArt-$\Sigma$~\citep{chen2024pixartsigma} &  6.34 & 27.62  & 0.54 \\
        Playground v2.5~\citep{li2024playground} &  6.84 & 29.39  & 0.56 \\
        SD3-medium~\citep{esser2024sd3}  & 11.92 & 27.83  & 0.62 \\
        DALL-E 3~\citep{dalle3}          &    -- &    --  & 0.67 \\
        FLUX-dev~\citep{FLUX2024}        & 10.15 & 27.47  & 0.67 \\
        FLUX-schnell~\citep{FLUX2024}    &  7.94 & 28.14  & 0.71 \\
        \hline & \\[-2.0ex]
        DiT (Ours, baseline)      &  6.90 & 27.37  & 0.63 \\
        \ours{} (Ours)  &  6.52 & 27.26  & 0.58 \\
        \hline \hline
      \end{tabular}
        }
  \end{center}
  \label{tab:eval_1k}
  \vspace{-5pt}
\end{table}

\begin{table*}[t]
  \caption{Comparison with existing text-to-video models on VBench~\citep{huang2024vbench}. The exact resolutions of our 180p and 360p videos are 192$\times$320 and 384$\times$640, respectively. We choose 12 evaluation dimensions. Higher metric values indicate better performance.}  
  \vspace{-5pt}
  \begin{center}
  \resizebox{\textwidth}{!}{
  \begin{tabular}{lcccccccccccc}
    \hline \hline & \\[-2.0ex]
    Models & \begin{tabular}{@{}c@{}}Subject \\ Consistency\end{tabular} & \begin{tabular}{@{}c@{}}Background \\ Consistency\end{tabular} & \begin{tabular}{@{}c@{}}Aesthetic \\ Quality\end{tabular} & \begin{tabular}{@{}c@{}}Object \\ Class\end{tabular} & Scene & \begin{tabular}{@{}c@{}}Human \\ Action\end{tabular} & \begin{tabular}{@{}c@{}}Apperance \\ Style\end{tabular} & \begin{tabular}{@{}c@{}}Temporal \\ Style \end{tabular} & \begin{tabular}{@{}c@{}}Temporal \\ Flickering\end{tabular} & \begin{tabular}{@{}c@{}}Motion \\ Smoothness \end{tabular} & \begin{tabular}{@{}c@{}}Dynamic \\ Degree\end{tabular} & \begin{tabular}{@{}c@{}}Overall \\ Consistency\end{tabular} \\ & \\[-2.0ex]
    \hline \hline & \\[-2.0ex]
    ModelScope       & 89.87 & 95.29 & 56.39 & 82.25 & 39.26 & 92.40 & 23.39 & 25.37 & 98.28 & 95.79 & 66.39 & 25.67 \\    
    LaVie            & 91.41 & 97.47 & 54.94 & 91.82 & 52.69 & 96.80 & 23.56 & 25.93 & 98.30 & 96.38 & 49.72 & 26.41 \\
    Show-1           & 95.53 & 98.02 & 57.35 & 93.07 & 47.03 & 95.60 & 23.06 & 25.28 & 99.12 & 98.24 & 44.44 & 27.46 \\
    OpenSora-v1.2    & 96.75 & 97.61 & 56.18 & 82.22 & 42.44 & 91.20 & 23.95 & 24.55 & 99.47 & 98.50 & 42.39 & 26.85 \\
    VideoCrafter-2.0 & 96.85 & 98.22 & 63.13 & 92.55 & 55.29 & 95.00 & 25.13 & 25.84 & 98.41 & 97.73 & 42.50 & 28.23 \\
    T2V-Turbo (VC2)  & 96.28 & 97.02 & 63.04 & 93.96 & 55.58 & 95.20 & 24.42 & 25.51 & 97.48 & 97.34 & 49.17 & 28.16 \\
    CogVideoX-5B     & 96.23 & 96.52 & 61.98 & 85.23 & 53.20 & 99.40 & 24.91 & 25.38 & 98.66 & 96.92 & 70.97 & 28.23 \\
    Pika             & 96.94 & 97.36 & 62.04 & 88.72 & 49.83 & 86.20 & 22.26 & 24.22 & 99.74 & 99.50 & 47.50 & 25.94 \\
    Kling (2024-07)  & 98.33 & 97.60 & 61.21 & 87.24 & 50.86 & 93.40 & 19.62 & 24.17 & 99.30 & 99.40 & 46.94 & 26.42 \\
    Gen3             & 97.10 & 96.62 & 63.34 & 87.81 & 54.57 & 96.40 & 24.31 & 25.33 & 98.61 & 99.23 & 60.14 & 26.69 \\
    ARLON            & 93.41 & 97.10 & 61.01 & 89.80 & 54.43 & --    & --    & --    & 99.37 & 98.92 & 52.77 & 27.27 \\
    Emu3             & 95.32 & 97.69 & 59.64 & 86.17 & 37.11 & 77.71 & 20.92 & 23.26 & 98.57 & 98.93 & 79.27 & 24.79 \\
    \hline & \\[-2.0ex]
    \ours{}-180p (Ours)   & 95.71 & 98.53 & 55.58 & 91.61 & 50.12 & 97.60 & 25.03 & 25.34 & 98.92 & 98.87 & 79.72 & 27.36 \\
    \ours{}-360p (Ours)   & 91.36 & 95.34 & 62.71 & 86.08 & 46.28 & 96.80 & 25.09 & 25.10 & 97.00 & 98.75 & 84.44 & 27.41 \\
    \hline \hline
  \end{tabular}}
  \end{center}
  \label{tab:eval_vbench}
  \vspace{-5pt}
\end{table*}

\subsubsection{Benchmark Evaluation}
We compare our generated images and videos with existing approaches on the widely applied benchmarks. Specifically, we evaluate 256$\times$256 images produced by our base image model on MSCOCO-30K~\citep{lin2014mscoco}, 1024$\times$1024 images produced by our 1K-tuned image model on MJHQ-30K~\citep{li2024playground} and GenEval~\citep{ghosh2024geneval}, as well as the videos produced by our 180p and 360p T2V models on VBench~\citep{huang2024vbench}, respectively. Our model follows the standard guidelines when generating the results without tricks like prompt rewriting. All image models we compared with are diffusion-based models.

\paragraph{256$\times$256 \& 1024$\times$1024 T2I.} 
As shown in Table~\ref{tab:eval_256p} and Table~\ref{tab:eval_1k}, we compare our generation results against a wide range of recent approaches. Our \ours{} model demonstrates strong performance, achieving results highly comparable to other state-of-the-art methods in terms of FID~\citep{heusel2017fid} and CLIP-Score~\citep{radford2021clip} when generating high-resolution images. Notably, our model, built on the DiT framework, surpasses DiT on both the MSCOCO and MJHQ benchmarks.
However, our model underperforms on GenEval, an object-centric evaluation metric. Upon analyzing the evaluation dimensions and text-image pairs, we observe that compared to the DiT baseline, the SSM-based solution exhibits a higher failure rate in generating structure-coherent objects and more struggles with creating co-occurring objects that align with the text prompt (score 0.67 \textit{vs.} 0.75). We hypothesize that this limitation stems from the inherent constraints of SSM-dense models, particularly their insufficient capacity to model long-range global dependencies effectively.

\paragraph{180p \& 360p T2V.} We compare our generated videos with ModelScope~\citep{wang2023modelscope}, LaVie~\citep{wang2023lavie}, Show-1~\citep{zhang2024show1}, OpenSora-v1.2~\citep{opensora2024v1p2}, VideoCrafter-2.0~\citep{chen2023videocrafter1}, T2V-Turbo (VC2)~\citep{li2024t2vturbo}, CogVideoX-5B~\citep{yang2024cogvideox}, Pika~\citep{pika2023}, Kling~\citep{kling2024}, Gen3~\citep{gen3_2024}, ARLON~\citep{li2024arlon}, and Emu3~\citep{wang2024emu3} on the VBench benchmark~\citep{huang2024vbench}. Emu3 is the only autoregressive model on the list, while all other approaches are diffusion-based. Our \ours{} model demonstrates highly comparable performance to existing methods across all evaluation dimensions. In particular, the 360p model excels in \textit{Aesthetic Quality}, as well as \textit{Appearance and Temporal Style}. Additionally, it generates accurate and realistic \textit{Human Actions} while producing videos with a significantly higher \textit{Dynamic Degree} and a good \textit{Overall Consistency}.

\subsubsection{Efficiency}

\begin{figure}[t]
  \centering
  \includegraphics[width=0.95\columnwidth]{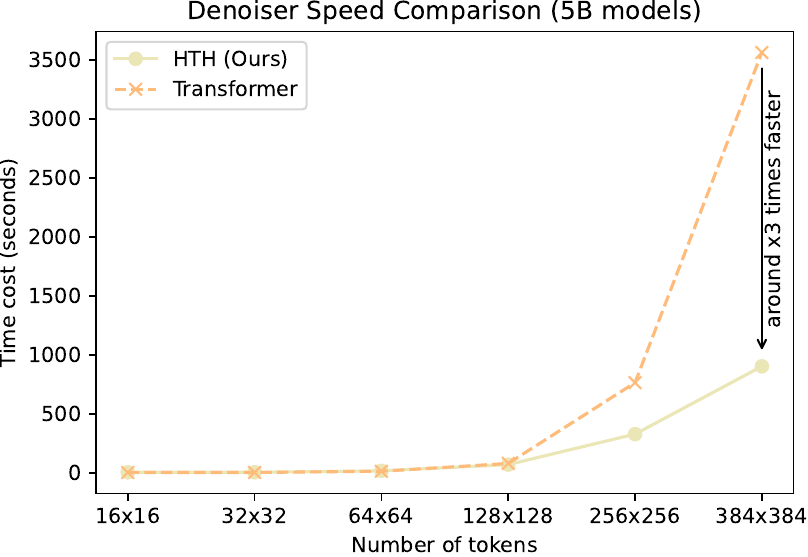}
  \vspace{-7pt}
  \caption{Denoiser inference speed comparison between HTH and Transformer (our DiT baseline).}
  \label{fig:speed_curve}
  \vspace{-10pt}
\end{figure}

In Figure~\ref{fig:speed_curve}, we compare the wall-time inference speed between a Transformer-based denoiser and our proposed HTH denoiser for processing varying numbers of tokens. Both models have 5B parameters and execute 50 (denoising) steps, with all self-attention layers employing the Flash-Attention 2 implementation~\citep{dao2023flashattention2}.
At a token length of 128$^{2}$, corresponding to the largest 2K images we tested with our HTH model, the speed difference is minor, with HTH being only 10 seconds faster. However, as the resolution increases, the performance gap widens exponentially. This observation is consistent with the quadratic computational complexity associated with global self-attention. A token length of 384$^{2}$ roughly corresponds to an 8K image or a 6-second 1K video at 24 FPS under our $\times$8 spatial and $\frac{16}{5}\times$ temporal compression rates and a patch size of 2; this suggests that the efficiency benefits of the SSM-based model become more pronounced for very long visual sequences.
The current implementation of bidirectional SSMs, like Hydra, presents significant potential for hardware (CUDA) optimization, including kernel operation integration. We intend to pursue these optimizations in future work.

\section{Conclusion}
\label{sec:conclusion}

In this paper, we propose the Hydra-Transformer Hybrid (\ours{}) diffusion model, a large-scale framework designed to push the boundaries of applying state space models (SSMs) to high-resolution image and long video generation. Our experiments show that using simple scanning strategies, models built with sequential and subquadratic token mixers, such as bidirectional SSMs, can achieve performance comparable to Transformer-based models while demonstrating superior inference efficiency and the potential to zero-shot higher-resolution images. We identify essential limitations and future directions of SSM-based generative models and share them in Appendix~\ref{appsec:limitations}.
We hope our findings will provide valuable insights and inspire future research on leveraging state space models and broader, more efficient sub-quadratic complexity architectures to tackle complex visual generation problems.

\bibliography{main_paper}

\begin{thebibliography}{104}
\providecommand{\natexlab}[1]{#1}
\providecommand{\url}[1]{\texttt{#1}}
\expandafter\ifx\csname urlstyle\endcsname\relax
  \providecommand{\doi}[1]{doi: #1}\else
  \providecommand{\doi}{doi: \begingroup \urlstyle{rm}\Url}\fi

\bibitem[Balaji et~al.(2022)Balaji, Nah, Huang, Vahdat, Song, Zhang, Kreis, Aittala, Aila, Laine, et~al.]{balaji2022ediff}
Balaji, Y., Nah, S., Huang, X., Vahdat, A., Song, J., Zhang, Q., Kreis, K., Aittala, M., Aila, T., Laine, S., et~al.
\newblock ediff-i: Text-to-image diffusion models with an ensemble of expert denoisers.
\newblock \emph{arXiv preprint arXiv:2211.01324}, 2022.

\bibitem[Bar-Tal et~al.(2024)Bar-Tal, Chefer, Tov, Herrmann, Paiss, Zada, Ephrat, Hur, Li, Michaeli, et~al.]{bar2024lumiere}
Bar-Tal, O., Chefer, H., Tov, O., Herrmann, C., Paiss, R., Zada, S., Ephrat, A., Hur, J., Li, Y., Michaeli, T., et~al.
\newblock Lumiere: A space-time diffusion model for video generation.
\newblock \emph{arXiv preprint arXiv:2401.12945}, 2024.

\bibitem[Beltagy et~al.(2020)Beltagy, Peters, and Cohan]{beltagy2020longformer}
Beltagy, I., Peters, M.~E., and Cohan, A.
\newblock Longformer: The long-document transformer.
\newblock \emph{arXiv preprint arXiv:2004.05150}, 2020.

\bibitem[Bick et~al.(2024)Bick, Li, Xing, Kolter, and Gu]{bick2024mohawk}
Bick, A., Li, K.~Y., Xing, E.~P., Kolter, J.~Z., and Gu, A.
\newblock Transformers to ssms: Distilling quadratic knowledge to subquadratic models.
\newblock \emph{arXiv preprint arXiv:2408.10189}, 2024.

\bibitem[{Black Forest Labs}(2024)]{FLUX2024}
{Black Forest Labs}.
\newblock Flux by black forest labs, 2024.
\newblock URL \url{https://github.com/black-forest-labs/flux}.

\bibitem[Blattmann et~al.(2023)Blattmann, Rombach, Ling, Dockhorn, Kim, Fidler, and Kreis]{blattmann2023align}
Blattmann, A., Rombach, R., Ling, H., Dockhorn, T., Kim, S.~W., Fidler, S., and Kreis, K.
\newblock Align your latents: High-resolution video synthesis with latent diffusion models.
\newblock In \emph{Proceedings of the IEEE/CVF Conference on Computer Vision and Pattern Recognition}, pp.\  22563--22575, 2023.

\bibitem[Botev et~al.(2024)Botev, De, Smith, Fernando, Muraru, Haroun, Berrada, Pascanu, Sessa, Dadashi, et~al.]{botev2024recurrentgemma}
Botev, A., De, S., Smith, S.~L., Fernando, A., Muraru, G.-C., Haroun, R., Berrada, L., Pascanu, R., Sessa, P.~G., Dadashi, R., et~al.
\newblock Recurrentgemma: Moving past transformers for efficient open language models.
\newblock \emph{arXiv preprint arXiv:2404.07839}, 2024.

\bibitem[Chang et~al.(2023)Chang, Zhang, Barber, Maschinot, Lezama, Jiang, Yang, Murphy, Freeman, Rubinstein, Li, and Krishnan]{chang2023muse}
Chang, H., Zhang, H., Barber, J., Maschinot, A., Lezama, J., Jiang, L., Yang, M.-H., Murphy, K., Freeman, W.~T., Rubinstein, M., Li, Y., and Krishnan, D.
\newblock Muse: Text-to-image generation via masked generative transformers.
\newblock \emph{arXiv preprint arXiv:2301.00704}, 2023.

\bibitem[Cheang et~al.(2024)Cheang, Chen, Jing, Kong, Li, Li, Liu, Wu, Xu, Yang, et~al.]{cheang2024gr2}
Cheang, C.-L., Chen, G., Jing, Y., Kong, T., Li, H., Li, Y., Liu, Y., Wu, H., Xu, J., Yang, Y., et~al.
\newblock Gr-2: A generative video-language-action model with web-scale knowledge for robot manipulation.
\newblock \emph{arXiv preprint arXiv:2410.06158}, 2024.

\bibitem[Chen et~al.(2023)Chen, Xia, He, Zhang, Cun, Yang, Xing, Liu, Chen, Wang, et~al.]{chen2023videocrafter1}
Chen, H., Xia, M., He, Y., Zhang, Y., Cun, X., Yang, S., Xing, J., Liu, Y., Chen, Q., Wang, X., et~al.
\newblock Videocrafter1: Open diffusion models for high-quality video generation.
\newblock \emph{arXiv preprint arXiv:2310.19512}, 2023.

\bibitem[Chen et~al.(2024{\natexlab{a}})Chen, Ge, Xie, Wu, Yao, Ren, Wang, Luo, Lu, and Li]{chen2024pixartsigma}
Chen, J., Ge, C., Xie, E., Wu, Y., Yao, L., Ren, X., Wang, Z., Luo, P., Lu, H., and Li, Z.
\newblock Pixart-$\backslash$sigma: Weak-to-strong training of diffusion transformer for 4k text-to-image generation.
\newblock \emph{arXiv preprint arXiv:2403.04692}, 2024{\natexlab{a}}.

\bibitem[Chen et~al.(2024{\natexlab{b}})Chen, Niu, Lu, Meng, and Zhou]{chen2024maskmamba}
Chen, W., Niu, L., Lu, Z., Meng, F., and Zhou, J.
\newblock Maskmamba: A hybrid mamba-transformer model for masked image generation.
\newblock \emph{arXiv preprint arXiv:2409.19937}, 2024{\natexlab{b}}.

\bibitem[Choromanski et~al.(2020)Choromanski, Likhosherstov, Dohan, Song, Gane, Sarlos, Hawkins, Davis, Mohiuddin, Kaiser, et~al.]{choromanski2020performers}
Choromanski, K., Likhosherstov, V., Dohan, D., Song, X., Gane, A., Sarlos, T., Hawkins, P., Davis, J., Mohiuddin, A., Kaiser, L., et~al.
\newblock Rethinking attention with performers.
\newblock \emph{arXiv preprint arXiv:2009.14794}, 2020.

\bibitem[Chung et~al.(2024)Chung, Hou, Longpre, Zoph, Tay, Fedus, Li, Wang, Dehghani, Brahma, et~al.]{chung2024scalingt5}
Chung, H.~W., Hou, L., Longpre, S., Zoph, B., Tay, Y., Fedus, W., Li, Y., Wang, X., Dehghani, M., Brahma, S., et~al.
\newblock Scaling instruction-finetuned language models.
\newblock \emph{Journal of Machine Learning Research}, 25\penalty0 (70):\penalty0 1--53, 2024.

\bibitem[Dao(2023)]{dao2023flashattention2}
Dao, T.
\newblock Flashattention-2: Faster attention with better parallelism and work partitioning.
\newblock \emph{arXiv preprint arXiv:2307.08691}, 2023.

\bibitem[Dao \& Gu(2024)Dao and Gu]{dao2024mamba2}
Dao, T. and Gu, A.
\newblock Transformers are ssms: Generalized models and efficient algorithms through structured state space duality.
\newblock \emph{arXiv preprint arXiv:2405.21060}, 2024.

\bibitem[De et~al.(2024)De, Smith, Fernando, Botev, Cristian-Muraru, Gu, Haroun, Berrada, Chen, Srinivasan, et~al.]{de2024griffin}
De, S., Smith, S.~L., Fernando, A., Botev, A., Cristian-Muraru, G., Gu, A., Haroun, R., Berrada, L., Chen, Y., Srinivasan, S., et~al.
\newblock Griffin: Mixing gated linear recurrences with local attention for efficient language models.
\newblock \emph{arXiv preprint arXiv:2402.19427}, 2024.

\bibitem[Dehghani et~al.(2023)Dehghani, Djolonga, Mustafa, Padlewski, Heek, Gilmer, Steiner, Caron, Geirhos, Alabdulmohsin, et~al.]{dehghani2023scaling}
Dehghani, M., Djolonga, J., Mustafa, B., Padlewski, P., Heek, J., Gilmer, J., Steiner, A.~P., Caron, M., Geirhos, R., Alabdulmohsin, I., et~al.
\newblock Scaling vision transformers to 22 billion parameters.
\newblock In \emph{International Conference on Machine Learning}, pp.\  7480--7512. PMLR, 2023.

\bibitem[Dhariwal \& Nichol(2021)Dhariwal and Nichol]{dhariwal2021diffusion}
Dhariwal, P. and Nichol, A.
\newblock Diffusion models beat gans on image synthesis.
\newblock \emph{Advances in neural information processing systems}, 34:\penalty0 8780--8794, 2021.

\bibitem[Dosovitskiy(2020)]{dosovitskiy2020vit}
Dosovitskiy, A.
\newblock An image is worth 16x16 words: Transformers for image recognition at scale.
\newblock \emph{arXiv preprint arXiv:2010.11929}, 2020.

\bibitem[Du et~al.(2024)Du, Yang, Dai, Dai, Nachum, Tenenbaum, Schuurmans, and Abbeel]{du2024learning}
Du, Y., Yang, S., Dai, B., Dai, H., Nachum, O., Tenenbaum, J., Schuurmans, D., and Abbeel, P.
\newblock Learning universal policies via text-guided video generation.
\newblock \emph{Advances in Neural Information Processing Systems}, 36, 2024.

\bibitem[Esser et~al.(2021)Esser, Rombach, and Ommer]{esser2021taming}
Esser, P., Rombach, R., and Ommer, B.
\newblock Taming transformers for high-resolution image synthesis.
\newblock In \emph{Proceedings of the IEEE/CVF conference on computer vision and pattern recognition}, pp.\  12873--12883, 2021.

\bibitem[Esser et~al.(2024)Esser, Kulal, Blattmann, Entezari, M{\"u}ller, Saini, Levi, Lorenz, Sauer, Boesel, et~al.]{esser2024sd3}
Esser, P., Kulal, S., Blattmann, A., Entezari, R., M{\"u}ller, J., Saini, H., Levi, Y., Lorenz, D., Sauer, A., Boesel, F., et~al.
\newblock Scaling rectified flow transformers for high-resolution image synthesis.
\newblock In \emph{Forty-first International Conference on Machine Learning}, 2024.

\bibitem[Fei et~al.(2024)Fei, Fan, Yu, Li, Zhang, and Huang]{fei2024dimba}
Fei, Z., Fan, M., Yu, C., Li, D., Zhang, Y., and Huang, J.
\newblock Dimba: Transformer-mamba diffusion models.
\newblock \emph{arXiv preprint arXiv:2406.01159}, 2024.

\bibitem[Gao et~al.(2024)Gao, Huang, Sun, Jie, Zhong, and Ma]{gao2024matten}
Gao, Y., Huang, J., Sun, X., Jie, Z., Zhong, Y., and Ma, L.
\newblock Matten: Video generation with mamba-attention.
\newblock \emph{arXiv preprint arXiv:2405.03025}, 2024.

\bibitem[Ghosh et~al.(2024)Ghosh, Hajishirzi, and Schmidt]{ghosh2024geneval}
Ghosh, D., Hajishirzi, H., and Schmidt, L.
\newblock Geneval: An object-focused framework for evaluating text-to-image alignment.
\newblock \emph{Advances in Neural Information Processing Systems}, 36, 2024.

\bibitem[Girdhar et~al.(2023)Girdhar, Singh, Brown, Duval, Azadi, Rambhatla, Shah, Yin, Parikh, and Misra]{girdhar2023emuvid}
Girdhar, R., Singh, M., Brown, A., Duval, Q., Azadi, S., Rambhatla, S.~S., Shah, A., Yin, X., Parikh, D., and Misra, I.
\newblock Emu video: Factorizing text-to-video generation by explicit image conditioning.
\newblock \emph{arXiv preprint arXiv:2311.10709}, 2023.

\bibitem[Gu \& Dao(2023)Gu and Dao]{gu2023mamba}
Gu, A. and Dao, T.
\newblock Mamba: Linear-time sequence modeling with selective state spaces.
\newblock \emph{arXiv preprint arXiv:2312.00752}, 2023.

\bibitem[Henry et~al.(2020)Henry, Dachapally, Pawar, and Chen]{henry2020qknorm}
Henry, A., Dachapally, P.~R., Pawar, S., and Chen, Y.
\newblock Query-key normalization for transformers.
\newblock \emph{arXiv preprint arXiv:2010.04245}, 2020.

\bibitem[Heusel et~al.(2017)Heusel, Ramsauer, Unterthiner, Nessler, and Hochreiter]{heusel2017fid}
Heusel, M., Ramsauer, H., Unterthiner, T., Nessler, B., and Hochreiter, S.
\newblock Gans trained by a two time-scale update rule converge to a local nash equilibrium.
\newblock \emph{Advances in neural information processing systems}, 30, 2017.

\bibitem[Ho \& Salimans(2022)Ho and Salimans]{ho2022classifier}
Ho, J. and Salimans, T.
\newblock Classifier-free diffusion guidance.
\newblock \emph{arXiv preprint arXiv:2207.12598}, 2022.

\bibitem[Ho et~al.(2020)Ho, Jain, and Abbeel]{ho2020ddpm}
Ho, J., Jain, A., and Abbeel, P.
\newblock Denoising diffusion probabilistic models.
\newblock \emph{Advances in neural information processing systems}, 33:\penalty0 6840--6851, 2020.

\bibitem[Ho et~al.(2022)Ho, Chan, Saharia, Whang, Gao, Gritsenko, Kingma, Poole, Norouzi, Fleet, and Salimans]{ho2022imagenvideo}
Ho, J., Chan, W., Saharia, C., Whang, J., Gao, R., Gritsenko, A., Kingma, D.~P., Poole, B., Norouzi, M., Fleet, D.~J., and Salimans, T.
\newblock Imagen video: High definition video generation with diffusion models.
\newblock \emph{arXiv preprint arXiv:2210.02303}, 2022.

\bibitem[Hong et~al.(2022)Hong, Ding, Zheng, Liu, and Tang]{hong2022cogvideo}
Hong, W., Ding, M., Zheng, W., Liu, X., and Tang, J.
\newblock Cogvideo: Large-scale pretraining for text-to-video generation via transformers.
\newblock \emph{arXiv preprint arXiv:2205.15868}, 2022.

\bibitem[Hong et~al.(2023)Hong, Zhang, Gu, Bi, Zhou, Liu, Liu, Sunkavalli, Bui, and Tan]{hong2023lrm}
Hong, Y., Zhang, K., Gu, J., Bi, S., Zhou, Y., Liu, D., Liu, F., Sunkavalli, K., Bui, T., and Tan, H.
\newblock Lrm: Large reconstruction model for single image to 3d.
\newblock \emph{arXiv preprint arXiv:2311.04400}, 2023.

\bibitem[Hu et~al.(2024)Hu, Baumann, Gui, Grebenkova, Ma, Fischer, and Ommer]{hu2024zigma}
Hu, V.~T., Baumann, S.~A., Gui, M., Grebenkova, O., Ma, P., Fischer, J., and Ommer, B.
\newblock Zigma: A dit-style zigzag mamba diffusion model.
\newblock \emph{arXiv preprint arXiv:2403.13802}, 2024.

\bibitem[Huang et~al.(2024)Huang, He, Yu, Zhang, Si, Jiang, Zhang, Wu, Jin, Chanpaisit, et~al.]{huang2024vbench}
Huang, Z., He, Y., Yu, J., Zhang, F., Si, C., Jiang, Y., Zhang, Y., Wu, T., Jin, Q., Chanpaisit, N., et~al.
\newblock Vbench: Comprehensive benchmark suite for video generative models.
\newblock In \emph{Proceedings of the IEEE/CVF Conference on Computer Vision and Pattern Recognition}, pp.\  21807--21818, 2024.

\bibitem[Hwang et~al.(2024)Hwang, Lahoti, Dao, and Gu]{hwang2024hydra}
Hwang, S., Lahoti, A., Dao, T., and Gu, A.
\newblock Hydra: Bidirectional state space models through generalized matrix mixers.
\newblock \emph{arXiv preprint arXiv:2407.09941}, 2024.

\bibitem[Jelassi et~al.(2024)Jelassi, Brandfonbrener, Kakade, and Malach]{jelassi2024repeat}
Jelassi, S., Brandfonbrener, D., Kakade, S.~M., and Malach, E.
\newblock Repeat after me: Transformers are better than state space models at copying.
\newblock \emph{arXiv preprint arXiv:2402.01032}, 2024.

\bibitem[Katharopoulos et~al.(2020)Katharopoulos, Vyas, Pappas, and Fleuret]{katharopoulos2020transformers}
Katharopoulos, A., Vyas, A., Pappas, N., and Fleuret, F.
\newblock Transformers are rnns: Fast autoregressive transformers with linear attention.
\newblock In \emph{International conference on machine learning}, pp.\  5156--5165. PMLR, 2020.

\bibitem[Kitaev et~al.(2020)Kitaev, Kaiser, and Levskaya]{kitaev2020reformer}
Kitaev, N., Kaiser, {\L}., and Levskaya, A.
\newblock Reformer: The efficient transformer.
\newblock \emph{arXiv preprint arXiv:2001.04451}, 2020.

\bibitem[{Kling AI}(2024)]{kling2024}
{Kling AI}.
\newblock Kling, 2024.
\newblock URL \url{https://klingai.com/}.

\bibitem[Kondratyuk et~al.(2023)Kondratyuk, Yu, Gu, Lezama, Huang, Schindler, Hornung, Birodkar, Yan, Chiu, et~al.]{kondratyuk2023videopoet}
Kondratyuk, D., Yu, L., Gu, X., Lezama, J., Huang, J., Schindler, G., Hornung, R., Birodkar, V., Yan, J., Chiu, M.-C., et~al.
\newblock Videopoet: A large language model for zero-shot video generation.
\newblock \emph{arXiv preprint arXiv:2312.14125}, 2023.

\bibitem[Lei~Ba et~al.(2016)Lei~Ba, Kiros, and Hinton]{lei2016layernorm}
Lei~Ba, J., Kiros, J.~R., and Hinton, G.~E.
\newblock Layer normalization.
\newblock \emph{ArXiv e-prints}, pp.\  arXiv--1607, 2016.

\bibitem[Li et~al.(2024{\natexlab{a}})Li, Kamko, Akhgari, Sabet, Xu, and Doshi]{li2024playground}
Li, D., Kamko, A., Akhgari, E., Sabet, A., Xu, L., and Doshi, S.
\newblock Playground v2. 5: Three insights towards enhancing aesthetic quality in text-to-image generation.
\newblock \emph{arXiv preprint arXiv:2402.17245}, 2024{\natexlab{a}}.

\bibitem[Li et~al.(2024{\natexlab{b}})Li, Feng, Fu, Wang, Basu, Chen, and Wang]{li2024t2vturbo}
Li, J., Feng, W., Fu, T.-J., Wang, X., Basu, S., Chen, W., and Wang, W.~Y.
\newblock T2v-turbo: Breaking the quality bottleneck of video consistency model with mixed reward feedback.
\newblock \emph{arXiv preprint arXiv:2405.18750}, 2024{\natexlab{b}}.

\bibitem[Li et~al.(2023)Li, Li, Zhu, Yu, Zhao, Wan, You, Shi, and Lin]{li2023instant}
Li, S., Li, C., Zhu, W., Yu, B., Zhao, Y., Wan, C., You, H., Shi, H., and Lin, Y.
\newblock Instant-3d: Instant neural radiance field training towards on-device ar/vr 3d reconstruction.
\newblock In \emph{Proceedings of the 50th Annual International Symposium on Computer Architecture}, pp.\  1--13, 2023.

\bibitem[Li et~al.(2024{\natexlab{c}})Li, Hu, Liu, Zhou, Choi, Meng, Guo, Li, Ling, and Wei]{li2024arlon}
Li, Z., Hu, S., Liu, S., Zhou, L., Choi, J., Meng, L., Guo, X., Li, J., Ling, H., and Wei, F.
\newblock Arlon: Boosting diffusion transformers with autoregressive models for long video generation.
\newblock \emph{arXiv preprint arXiv:2410.20502}, 2024{\natexlab{c}}.

\bibitem[Lieber et~al.(2024)Lieber, Lenz, Bata, Cohen, Osin, Dalmedigos, Safahi, Meirom, Belinkov, Shalev-Shwartz, et~al.]{lieber2024jamba}
Lieber, O., Lenz, B., Bata, H., Cohen, G., Osin, J., Dalmedigos, I., Safahi, E., Meirom, S., Belinkov, Y., Shalev-Shwartz, S., et~al.
\newblock Jamba: A hybrid transformer-mamba language model.
\newblock \emph{arXiv preprint arXiv:2403.19887}, 2024.

\bibitem[Lin et~al.(2014)Lin, Maire, Belongie, Hays, Perona, Ramanan, Doll{\'a}r, and Zitnick]{lin2014mscoco}
Lin, T.-Y., Maire, M., Belongie, S., Hays, J., Perona, P., Ramanan, D., Doll{\'a}r, P., and Zitnick, C.~L.
\newblock Microsoft coco: Common objects in context.
\newblock In \emph{Computer Vision--ECCV 2014: 13th European Conference, Zurich, Switzerland, September 6-12, 2014, Proceedings, Part V 13}, pp.\  740--755. Springer, 2014.

\bibitem[Liu et~al.(2024)Liu, Yu, Tan, and Wang]{liu2024linfusion}
Liu, S., Yu, W., Tan, Z., and Wang, X.
\newblock Linfusion: 1 gpu, 1 minute, 16k image.
\newblock \emph{arXiv preprint arXiv:2409.02097}, 2024.

\bibitem[Liu et~al.(2023)Liu, Zhang, Ma, Peng, et~al.]{liu2023instaflow}
Liu, X., Zhang, X., Ma, J., Peng, J., et~al.
\newblock Instaflow: One step is enough for high-quality diffusion-based text-to-image generation.
\newblock In \emph{The Twelfth International Conference on Learning Representations}, 2023.

\bibitem[Merrill et~al.(2024)Merrill, Petty, and Sabharwal]{merrill2024illusion}
Merrill, W., Petty, J., and Sabharwal, A.
\newblock The illusion of state in state-space models.
\newblock \emph{arXiv preprint arXiv:2404.08819}, 2024.

\bibitem[{Open-Sora}(2024)]{opensora2024v1p2}
{Open-Sora}.
\newblock Open-sora 1.2 report, 2024.
\newblock URL \url{https://github.com/hpcaitech/Open-Sora}.

\bibitem[{OpenAI}(2023)]{dalle3}
{OpenAI}.
\newblock Dall-e 3 by openai, 2023.
\newblock URL \url{https://openai.com/dall-e-3}.

\bibitem[Peebles \& Xie(2023)Peebles and Xie]{peebles2023dit}
Peebles, W. and Xie, S.
\newblock Scalable diffusion models with transformers.
\newblock In \emph{Proceedings of the IEEE/CVF International Conference on Computer Vision}, pp.\  4195--4205, 2023.

\bibitem[Peng et~al.(2023)Peng, Alcaide, Anthony, Albalak, Arcadinho, Biderman, Cao, Cheng, Chung, Grella, et~al.]{peng2023rwkv}
Peng, B., Alcaide, E., Anthony, Q., Albalak, A., Arcadinho, S., Biderman, S., Cao, H., Cheng, X., Chung, M., Grella, M., et~al.
\newblock Rwkv: Reinventing rnns for the transformer era.
\newblock \emph{arXiv preprint arXiv:2305.13048}, 2023.

\bibitem[{Pika Labs}(2023)]{pika2023}
{Pika Labs}.
\newblock Pika, 2023.
\newblock URL \url{https://github.com/hpcaitech/Open-Sora}.

\bibitem[Podell et~al.(2023)Podell, English, Lacey, Blattmann, Dockhorn, M{\"u}ller, Penna, and Rombach]{podell2023sdxl}
Podell, D., English, Z., Lacey, K., Blattmann, A., Dockhorn, T., M{\"u}ller, J., Penna, J., and Rombach, R.
\newblock Sdxl: Improving latent diffusion models for high-resolution image synthesis.
\newblock \emph{arXiv preprint arXiv:2307.01952}, 2023.

\bibitem[Qin et~al.(2022)Qin, Sun, Deng, Li, Wei, Lv, Yan, Kong, and Zhong]{qin2022cosformer}
Qin, Z., Sun, W., Deng, H., Li, D., Wei, Y., Lv, B., Yan, J., Kong, L., and Zhong, Y.
\newblock cosformer: Rethinking softmax in attention.
\newblock \emph{arXiv preprint arXiv:2202.08791}, 2022.

\bibitem[Radford et~al.(2021)Radford, Kim, Hallacy, Ramesh, Goh, Agarwal, Sastry, Askell, Mishkin, Clark, et~al.]{radford2021clip}
Radford, A., Kim, J.~W., Hallacy, C., Ramesh, A., Goh, G., Agarwal, S., Sastry, G., Askell, A., Mishkin, P., Clark, J., et~al.
\newblock Learning transferable visual models from natural language supervision.
\newblock In \emph{International conference on machine learning}, pp.\  8748--8763. PMLR, 2021.

\bibitem[Raffel et~al.(2020)Raffel, Shazeer, Roberts, Lee, Narang, Matena, Zhou, Li, and Liu]{raffel2020exploringt5}
Raffel, C., Shazeer, N., Roberts, A., Lee, K., Narang, S., Matena, M., Zhou, Y., Li, W., and Liu, P.~J.
\newblock Exploring the limits of transfer learning with a unified text-to-text transformer.
\newblock \emph{Journal of machine learning research}, 21\penalty0 (140):\penalty0 1--67, 2020.

\bibitem[Ramesh et~al.(2022)Ramesh, Dhariwal, Nichol, Chu, and Chen]{ramesh2022dalle2}
Ramesh, A., Dhariwal, P., Nichol, A., Chu, C., and Chen, M.
\newblock Hierarchical text-conditional image generation with clip latents.
\newblock \emph{arXiv preprint arXiv:2204.06125}, 1\penalty0 (2):\penalty0 3, 2022.

\bibitem[Rombach et~al.(2022)Rombach, Blattmann, Lorenz, Esser, and Ommer]{rombach2022ldm}
Rombach, R., Blattmann, A., Lorenz, D., Esser, P., and Ommer, B.
\newblock High-resolution image synthesis with latent diffusion models.
\newblock In \emph{Proceedings of the IEEE/CVF conference on computer vision and pattern recognition}, pp.\  10684--10695, 2022.

\bibitem[Ruiz et~al.(2023)Ruiz, Li, Jampani, Pritch, Rubinstein, and Aberman]{ruiz2023dreambooth}
Ruiz, N., Li, Y., Jampani, V., Pritch, Y., Rubinstein, M., and Aberman, K.
\newblock Dreambooth: Fine tuning text-to-image diffusion models for subject-driven generation.
\newblock In \emph{Proceedings of the IEEE/CVF conference on computer vision and pattern recognition}, pp.\  22500--22510, 2023.

\bibitem[{Runway}(2024)]{gen3_2024}
{Runway}.
\newblock Introducing gen-3 alpha: A new frontier for video generation, 2024.
\newblock URL \url{https://runwayml.com/research/introducing-gen-3-alpha}.

\bibitem[Saharia et~al.(2022)Saharia, Chan, Saxena, Li, Whang, Denton, Ghasemipour, Ayan, Mahdavi, Lopes, Salimans, Ho, Fleet, and Norouzi]{saharia2022imagen}
Saharia, C., Chan, W., Saxena, S., Li, L., Whang, J., Denton, E., Ghasemipour, S. K.~S., Ayan, B.~K., Mahdavi, S.~S., Lopes, R.~G., Salimans, T., Ho, J., Fleet, D.~J., and Norouzi, M.
\newblock Photorealistic text-to-image diffusion models with deep language understanding.
\newblock \emph{arXiv preprint arXiv:2205.11487}, 2022.

\bibitem[Shen et~al.(2021)Shen, Zhang, Zhao, Yi, and Li]{shen2021efficient}
Shen, Z., Zhang, M., Zhao, H., Yi, S., and Li, H.
\newblock Efficient attention: Attention with linear complexities.
\newblock In \emph{Proceedings of the IEEE/CVF winter conference on applications of computer vision}, pp.\  3531--3539, 2021.

\bibitem[Shi et~al.(2024)Shi, Dong, Li, and Xu]{shi2024vssd}
Shi, Y., Dong, M., Li, M., and Xu, C.
\newblock Vssd: Vision mamba with non-causal state space duality.
\newblock \emph{arXiv preprint arXiv:2407.18559}, 2024.

\bibitem[Song et~al.(2020)Song, Meng, and Ermon]{song2020ddim}
Song, J., Meng, C., and Ermon, S.
\newblock Denoising diffusion implicit models.
\newblock \emph{arXiv preprint arXiv:2010.02502}, 2020.

\bibitem[{Stability AI}(2024)]{deepfloyd2024}
{Stability AI}.
\newblock If by deepfloyd lab at stability ai, 2024.
\newblock URL \url{https://stability.ai/news/deepfloyd-if-text-to-image-model}.

\bibitem[Su et~al.(2024)Su, Ahmed, Lu, Pan, Bo, and Liu]{su2024roformer}
Su, J., Ahmed, M., Lu, Y., Pan, S., Bo, W., and Liu, Y.
\newblock Roformer: Enhanced transformer with rotary position embedding.
\newblock \emph{Neurocomputing}, 568:\penalty0 127063, 2024.

\bibitem[Sun et~al.(2024)Sun, Jiang, Chen, Zhang, Peng, Luo, and Yuan]{sun2024autoregressive}
Sun, P., Jiang, Y., Chen, S., Zhang, S., Peng, B., Luo, P., and Yuan, Z.
\newblock Autoregressive model beats diffusion: Llama for scalable image generation.
\newblock \emph{arXiv preprint arXiv:2406.06525}, 2024.

\bibitem[Sun et~al.(2023)Sun, Dong, Huang, Ma, Xia, Xue, Wang, and Wei]{sun2023retentive}
Sun, Y., Dong, L., Huang, S., Ma, S., Xia, Y., Xue, J., Wang, J., and Wei, F.
\newblock Retentive network: A successor to transformer for large language models.
\newblock \emph{arXiv preprint arXiv:2307.08621}, 2023.

\bibitem[Vaswani et~al.(2017)Vaswani, Shazeer, Parmar, Uszkoreit, Jones, Gomez, Kaiser, and Polosukhin]{vaswani2017attention}
Vaswani, A., Shazeer, N., Parmar, N., Uszkoreit, J., Jones, L., Gomez, A.~N., Kaiser, L., and Polosukhin, I.
\newblock Attention is all you need.
\newblock \emph{Advances in Neural Information Processing Systems}, 2017.

\bibitem[Waleffe et~al.(2024)Waleffe, Byeon, Riach, Norick, Korthikanti, Dao, Gu, Hatamizadeh, Singh, Narayanan, et~al.]{waleffe2024empirical}
Waleffe, R., Byeon, W., Riach, D., Norick, B., Korthikanti, V., Dao, T., Gu, A., Hatamizadeh, A., Singh, S., Narayanan, D., et~al.
\newblock An empirical study of mamba-based language models.
\newblock \emph{arXiv preprint arXiv:2406.07887}, 2024.

\bibitem[Wang et~al.(2023{\natexlab{a}})Wang, Yuan, Chen, Zhang, Wang, and Zhang]{wang2023modelscope}
Wang, J., Yuan, H., Chen, D., Zhang, Y., Wang, X., and Zhang, S.
\newblock Modelscope text-to-video technical report.
\newblock \emph{arXiv preprint arXiv:2308.06571}, 2023{\natexlab{a}}.

\bibitem[Wang et~al.(2024{\natexlab{a}})Wang, Paliotta, May, Rush, and Dao]{wang2024mambainllama}
Wang, J., Paliotta, D., May, A., Rush, A.~M., and Dao, T.
\newblock The mamba in the llama: Distilling and accelerating hybrid models.
\newblock \emph{arXiv preprint arXiv:2408.15237}, 2024{\natexlab{a}}.

\bibitem[Wang et~al.(2020)Wang, Li, Khabsa, Fang, and Ma]{wang2020linformer}
Wang, S., Li, B.~Z., Khabsa, M., Fang, H., and Ma, H.
\newblock Linformer: Self-attention with linear complexity.
\newblock \emph{arXiv preprint arXiv:2006.04768}, 2020.

\bibitem[Wang et~al.(2024{\natexlab{b}})Wang, Leroy, Cabon, Chidlovskii, and Revaud]{wang2024dust3r}
Wang, S., Leroy, V., Cabon, Y., Chidlovskii, B., and Revaud, J.
\newblock Dust3r: Geometric 3d vision made easy.
\newblock In \emph{Proceedings of the IEEE/CVF Conference on Computer Vision and Pattern Recognition}, pp.\  20697--20709, 2024{\natexlab{b}}.

\bibitem[Wang et~al.(2024{\natexlab{c}})Wang, Zhang, Luo, Sun, Cui, Wang, Zhang, Wang, Li, Yu, et~al.]{wang2024emu3}
Wang, X., Zhang, X., Luo, Z., Sun, Q., Cui, Y., Wang, J., Zhang, F., Wang, Y., Li, Z., Yu, Q., et~al.
\newblock Emu3: Next-token prediction is all you need.
\newblock \emph{arXiv preprint arXiv:2409.18869}, 2024{\natexlab{c}}.

\bibitem[Wang et~al.(2023{\natexlab{b}})Wang, Chen, Ma, Zhou, Huang, Wang, Yang, He, Yu, Yang, et~al.]{wang2023lavie}
Wang, Y., Chen, X., Ma, X., Zhou, S., Huang, Z., Wang, Y., Yang, C., He, Y., Yu, J., Yang, P., et~al.
\newblock Lavie: High-quality video generation with cascaded latent diffusion models.
\newblock \emph{arXiv preprint arXiv:2309.15103}, 2023{\natexlab{b}}.

\bibitem[Watson et~al.(2024)Watson, Saxena, Li, Tagliasacchi, and Fleet]{watson2024controlling}
Watson, D., Saxena, S., Li, L., Tagliasacchi, A., and Fleet, D.~J.
\newblock Controlling space and time with diffusion models.
\newblock \emph{arXiv preprint arXiv:2407.07860}, 2024.

\bibitem[Wu et~al.(2023)Wu, Jing, Cheang, Chen, Xu, Li, Liu, Li, and Kong]{wu2023gr1}
Wu, H., Jing, Y., Cheang, C., Chen, G., Xu, J., Li, X., Liu, M., Li, H., and Kong, T.
\newblock Unleashing large-scale video generative pre-training for visual robot manipulation.
\newblock \emph{arXiv preprint arXiv:2312.13139}, 2023.

\bibitem[Xie et~al.(2024)Xie, Chen, Chen, Cai, Tang, Lin, Zhang, Li, Zhu, Lu, and Han]{xie2024sana}
Xie, E., Chen, J., Chen, J., Cai, H., Tang, H., Lin, Y., Zhang, Z., Li, M., Zhu, L., Lu, Y., and Han, S.
\newblock Sana: Efficient high-resolution image synthesis with linear diffusion transformers.
\newblock \emph{arXiv preprint arXiv:2410.10629}, 2024.

\bibitem[Xiong et~al.(2020)Xiong, Yang, He, Zheng, Zheng, Xing, Zhang, Lan, Wang, and Liu]{xiong2020onlayer}
Xiong, R., Yang, Y., He, D., Zheng, K., Zheng, S., Xing, C., Zhang, H., Lan, Y., Wang, L., and Liu, T.
\newblock On layer normalization in the transformer architecture.
\newblock In \emph{International Conference on Machine Learning}, pp.\  10524--10533. PMLR, 2020.

\bibitem[Xu et~al.(2024)Xu, Yang, Wang, Du, and Chen]{xu2024surveymamba}
Xu, R., Yang, S., Wang, Y., Du, B., and Chen, H.
\newblock A survey on vision mamba: Models, applications and challenges.
\newblock \emph{arXiv preprint arXiv:2404.18861}, 2024.

\bibitem[Xu et~al.(2023)Xu, Tan, Luan, Bi, Wang, Li, Shi, Sunkavalli, Wetzstein, Xu, et~al.]{xu2023dmv3d}
Xu, Y., Tan, H., Luan, F., Bi, S., Wang, P., Li, J., Shi, Z., Sunkavalli, K., Wetzstein, G., Xu, Z., et~al.
\newblock Dmv3d: Denoising multi-view diffusion using 3d large reconstruction model.
\newblock \emph{arXiv preprint arXiv:2311.09217}, 2023.

\bibitem[Xue et~al.(2024)Xue, Song, Guo, Liu, Zong, Liu, and Luo]{xue2024raphael}
Xue, Z., Song, G., Guo, Q., Liu, B., Zong, Z., Liu, Y., and Luo, P.
\newblock Raphael: Text-to-image generation via large mixture of diffusion paths.
\newblock \emph{Advances in Neural Information Processing Systems}, 36, 2024.

\bibitem[Yan et~al.(2023)Yan, Gu, and Rush]{yan2023difussm}
Yan, J.~N., Gu, J., and Rush, A.~M.
\newblock Diffusion models without attention.
\newblock \emph{arXiv preprint arXiv:2311.18257}, 2023.

\bibitem[Yan et~al.(2021)Yan, Zhang, Abbeel, and Srinivas]{yan2021videogpt}
Yan, W., Zhang, Y., Abbeel, P., and Srinivas, A.
\newblock Videogpt: Video generation using vq-vae and transformers.
\newblock \emph{arXiv preprint arXiv:2104.10157}, 2021.

\bibitem[Yang et~al.(2023{\natexlab{a}})Yang, Du, Ghasemipour, Tompson, Schuurmans, and Abbeel]{yang2023UniSim}
Yang, M., Du, Y., Ghasemipour, K., Tompson, J., Schuurmans, D., and Abbeel, P.
\newblock Learning interactive real-world simulators.
\newblock \emph{arXiv preprint arXiv:2310.06114}, 2023{\natexlab{a}}.

\bibitem[Yang et~al.(2023{\natexlab{b}})Yang, Wang, Shen, Panda, and Kim]{yang2023gated}
Yang, S., Wang, B., Shen, Y., Panda, R., and Kim, Y.
\newblock Gated linear attention transformers with hardware-efficient training.
\newblock \emph{arXiv preprint arXiv:2312.06635}, 2023{\natexlab{b}}.

\bibitem[Yang et~al.(2024)Yang, Teng, Zheng, Ding, Huang, Xu, Yang, Hong, Zhang, Feng, et~al.]{yang2024cogvideox}
Yang, Z., Teng, J., Zheng, W., Ding, M., Huang, S., Xu, J., Yang, Y., Hong, W., Zhang, X., Feng, G., et~al.
\newblock Cogvideox: Text-to-video diffusion models with an expert transformer.
\newblock \emph{arXiv preprint arXiv:2408.06072}, 2024.

\bibitem[Yi et~al.(2024)Yi, Wu, Shen, Xu, Zhou, Lim, Yan, Wang, and Zhang]{yi2024mvgamba}
Yi, X., Wu, Z., Shen, Q., Xu, Q., Zhou, P., Lim, J.-H., Yan, S., Wang, X., and Zhang, H.
\newblock Mvgamba: Unify 3d content generation as state space sequence modeling.
\newblock \emph{arXiv preprint arXiv:2406.06367}, 2024.

\bibitem[Yu et~al.(2023)Yu, Lezama, Gundavarapu, Versari, Sohn, Minnen, Cheng, Birodkar, Gupta, Gu, et~al.]{yu2023magvitv2}
Yu, L., Lezama, J., Gundavarapu, N.~B., Versari, L., Sohn, K., Minnen, D., Cheng, Y., Birodkar, V., Gupta, A., Gu, X., et~al.
\newblock Language model beats diffusion--tokenizer is key to visual generation.
\newblock \emph{arXiv preprint arXiv:2310.05737}, 2023.

\bibitem[Zhang \& Sennrich(2019)Zhang and Sennrich]{zhang2019rmsnorm}
Zhang, B. and Sennrich, R.
\newblock Root mean square layer normalization.
\newblock \emph{Advances in Neural Information Processing Systems}, 32, 2019.

\bibitem[Zhang et~al.(2024{\natexlab{a}})Zhang, Wu, Liu, Zhao, Ran, Gu, Gao, and Shou]{zhang2024show1}
Zhang, D.~J., Wu, J.~Z., Liu, J.-W., Zhao, R., Ran, L., Gu, Y., Gao, D., and Shou, M.~Z.
\newblock Show-1: Marrying pixel and latent diffusion models for text-to-video generation.
\newblock \emph{International Journal of Computer Vision}, pp.\  1--15, 2024{\natexlab{a}}.

\bibitem[Zhang et~al.(2024{\natexlab{b}})Zhang, Zhu, Wang, Zhang, Chen, Wang, and Ye]{zhang2024surveymamba}
Zhang, H., Zhu, Y., Wang, D., Zhang, L., Chen, T., Wang, Z., and Ye, Z.
\newblock A survey on visual mamba.
\newblock \emph{Applied Sciences}, 14\penalty0 (13):\penalty0 5683, 2024{\natexlab{b}}.

\bibitem[Zhang et~al.(2024{\natexlab{c}})Zhang, Liu, Reid, Hartley, Zhuang, and Tang]{zhang2024motion}
Zhang, Z., Liu, A., Reid, I., Hartley, R., Zhuang, B., and Tang, H.
\newblock Motion mamba: Efficient and long sequence motion generation with hierarchical and bidirectional selective ssm.
\newblock \emph{arXiv e-prints}, pp.\  arXiv--2403, 2024{\natexlab{c}}.

\bibitem[Zhou et~al.(2024)Zhou, Yu, Babu, Tirumala, Yasunaga, Shamis, Kahn, Ma, Zettlemoyer, and Levy]{zhou2024transfusion}
Zhou, C., Yu, L., Babu, A., Tirumala, K., Yasunaga, M., Shamis, L., Kahn, J., Ma, X., Zettlemoyer, L., and Levy, O.
\newblock Transfusion: Predict the next token and diffuse images with one multi-modal model.
\newblock \emph{arXiv preprint arXiv:2408.11039}, 2024.

\bibitem[Zhu et~al.(2024)Zhu, Huang, Liao, Liew, Yan, Feng, and Wang]{zhu2024dig}
Zhu, L., Huang, Z., Liao, B., Liew, J.~H., Yan, H., Feng, J., and Wang, X.
\newblock Dig: Scalable and efficient diffusion models with gated linear attention.
\newblock \emph{arXiv preprint arXiv:2405.18428}, 2024.

\bibitem[Ziwen et~al.(2024)Ziwen, Tan, Zhang, Bi, Luan, Hong, Fuxin, and Xu]{ziwen2024longlrm}
Ziwen, C., Tan, H., Zhang, K., Bi, S., Luan, F., Hong, Y., Fuxin, L., and Xu, Z.
\newblock Long-lrm: Long-sequence large reconstruction model for wide-coverage gaussian splats.
\newblock \emph{arXiv preprint arXiv:2410.12781}, 2024.

\bibitem[Zuo et~al.(2022)Zuo, Liu, Jiao, Charles, Manavoglu, Zhao, and Gao]{zuo2022augmented}
Zuo, S., Liu, X., Jiao, J., Charles, D., Manavoglu, E., Zhao, T., and Gao, J.
\newblock Efficient long sequence modeling via state space augmented transformer.
\newblock \emph{arXiv preprint arXiv:2212.08136}, 2022.

\end{thebibliography}
\bibliographystyle{icml2025}

\newpage
\appendix
\onecolumn
\section{Limitations and Future Directions}
\label{appsec:limitations}
Due to the extensive scope of this work and constraints in computational resources, we identify and share several limitations of our system that we believe are valuable for future research:
\begin{itemize}
    \item \textbf{SSM-friendly pipeline}. Our model is built upon the existing DiT~\citep{peebles2023dit} pipeline, which is specifically tailored and optimized for Transformer-based models and global attention operations. However, certain critical components, such as positional encoding, may be suboptimal for informing SSMs about spatial continuity. Improving these aspects is essential to better align the pipeline with SSM-based architectures.
    \item \textbf{Conditioning mechanisms}. The sequential nature of SSMs imposes constraints on how conditions are introduced to latent representations. In this work, we employ simple cross-attention to incorporate textual conditions, but the results generally exhibit inferior semantic understanding (\textit{e.g.}, worse entity composition and structural coherence) compared to Transformer-based models. Developing more effective conditioning mechanisms for SSMs is a promising direction for future exploration.  
    \item \textbf{SSM optimization}. The formulation and hardware optimization of SSMs remain at an early stage. Further research is needed to design SSMs better suited for processing N-dimensional visual data while maintaining high computational efficiency. 
\end{itemize}

\section{Additional Results}

\subsection{Comparison of Token Mixers}
\label{appsec:token_mixers}
\begin{figure*}[h]
  \centering
  \includegraphics[width=0.90\textwidth]{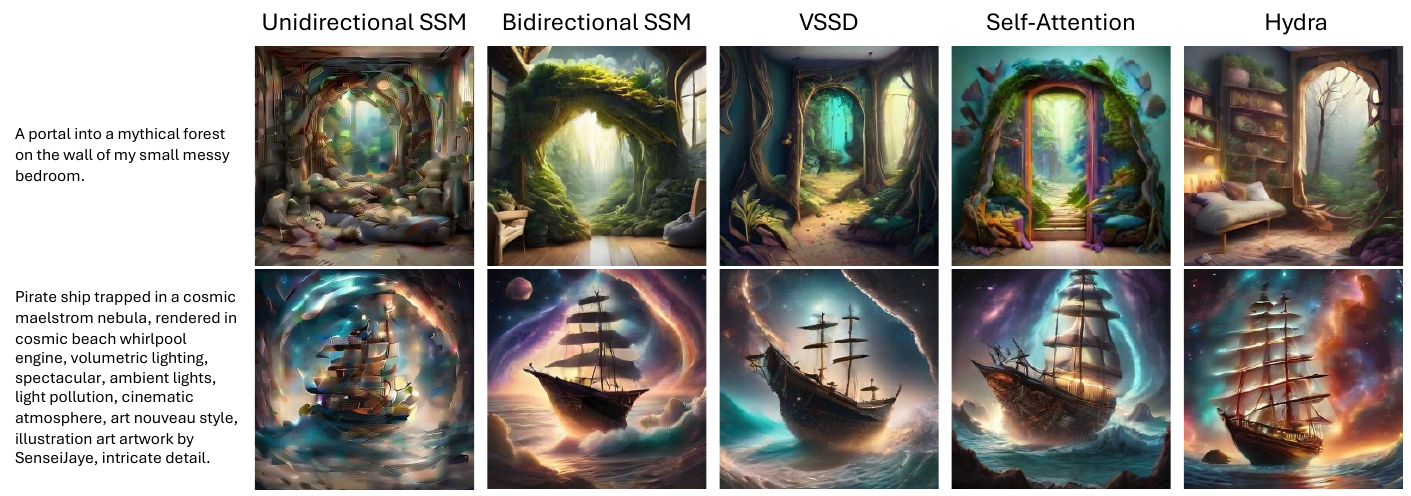}
  \vspace{-5pt}
  \caption{Example 256$\times$256-resolution T2I generation results of different token mixers. SSMs and Hydra models are based on Mamba-2~\citep{dao2024mamba2}. The Bidirectional SSM and Hydra apply interleaved horizontal and vertical raster scans across model blocks. None of the models in this small-scale experiment is hybrid (\textit{i.e.}, all blocks use the same token mixer).}
  \label{fig:scan_compare}
\end{figure*}

As shown in our small-scale experiment (Figure~\ref{fig:scan_compare}), we compared the generation quality across unidirectional SSM, bidirectional SSM, the state-sharing VSSD~\citep{shi2024vssd}, self-attention, and Hydra. We constructed 1B-parameter models with the different token mixers above while keeping all other components identical and trained each model for 300K iterations. Hydra demonstrated clearly superior visual quality than the other SSM-based models, particularly in structural coherence and correctness, while achieving better semantic alignment with the text prompt. Its performance is also highly comparable to the model employing self-attention.

\subsection{Higher-Res Zero-shot Generation}
\label{appsec:highres_zeroshot}
\begin{figure*}[h]
  \centering
  \includegraphics[width=0.75\textwidth]{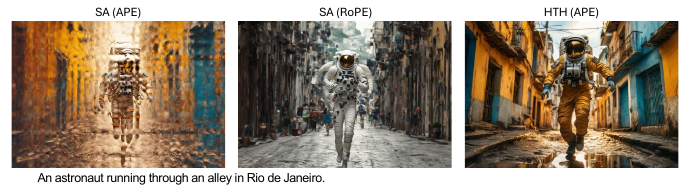}
  \vspace{-5pt}
  \caption{Zero-shot comparison at higher resolution. We compare our \ours{} model with the same-scale Transformer-based models (SA: Self-Attention) with Absolute Positional Embedding (APE) or Rotary Positional Embedding (RoPE). All models are trained on 256p image data and evaluated on 416$\times$608 images in a zero-shot manner.}
  \label{fig:zeroshot_compare}
\end{figure*}

We further compare the zero-shot performance of our SSM-dominant architecture with pure Transformer models, as shown in Figure~\ref{fig:zeroshot_compare}. All models are of the same scale and are trained exclusively on 256p images, but are tasked with generating 512p images. The Transformer-based model with absolute positional embedding (APE)~\citep{vaswani2017attention} struggles to generalize to higher resolutions, resulting in severe checkerboard artifacts. While models using relative positional encoding, such as RoPE~\citep{su2024roformer}, achieve better results, noticeable inconsistency and noise persist. In contrast, our proposed \ours{} model naturally generalizes to synthesize $\times$4 times larger images, benefiting from the strong locality properties inherent to the state space model.

\subsection{Visualization}
\label{appsec:visualization}
The following pages provide additional visualization of generated images in various aspect ratios up to 2560$\times$2560 resolution, and some extracted frames from generated 360p 128-frame videos. Please also refer to the Supplementary Materials for more playable videos.

\begin{figure*}[h!]
  \centering
  \includegraphics[width=0.95\textwidth]{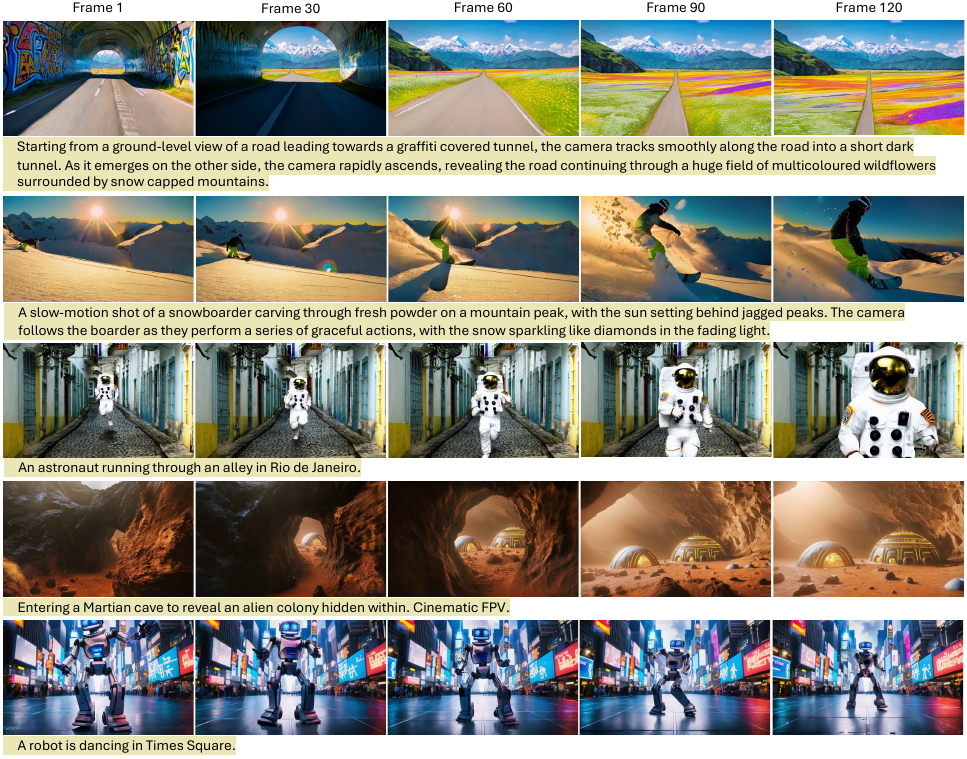}
  \vspace{-8pt}
  \caption{Text-to-360p 128 frames video generation results produced by our Hydra-Transformer Hybrid model. Please zoom in for a clearer visualization.}
  \label{fig:visualize_3}
\end{figure*}

\begin{figure*}[p]
  \centering
  \includegraphics[width=0.88\textwidth]{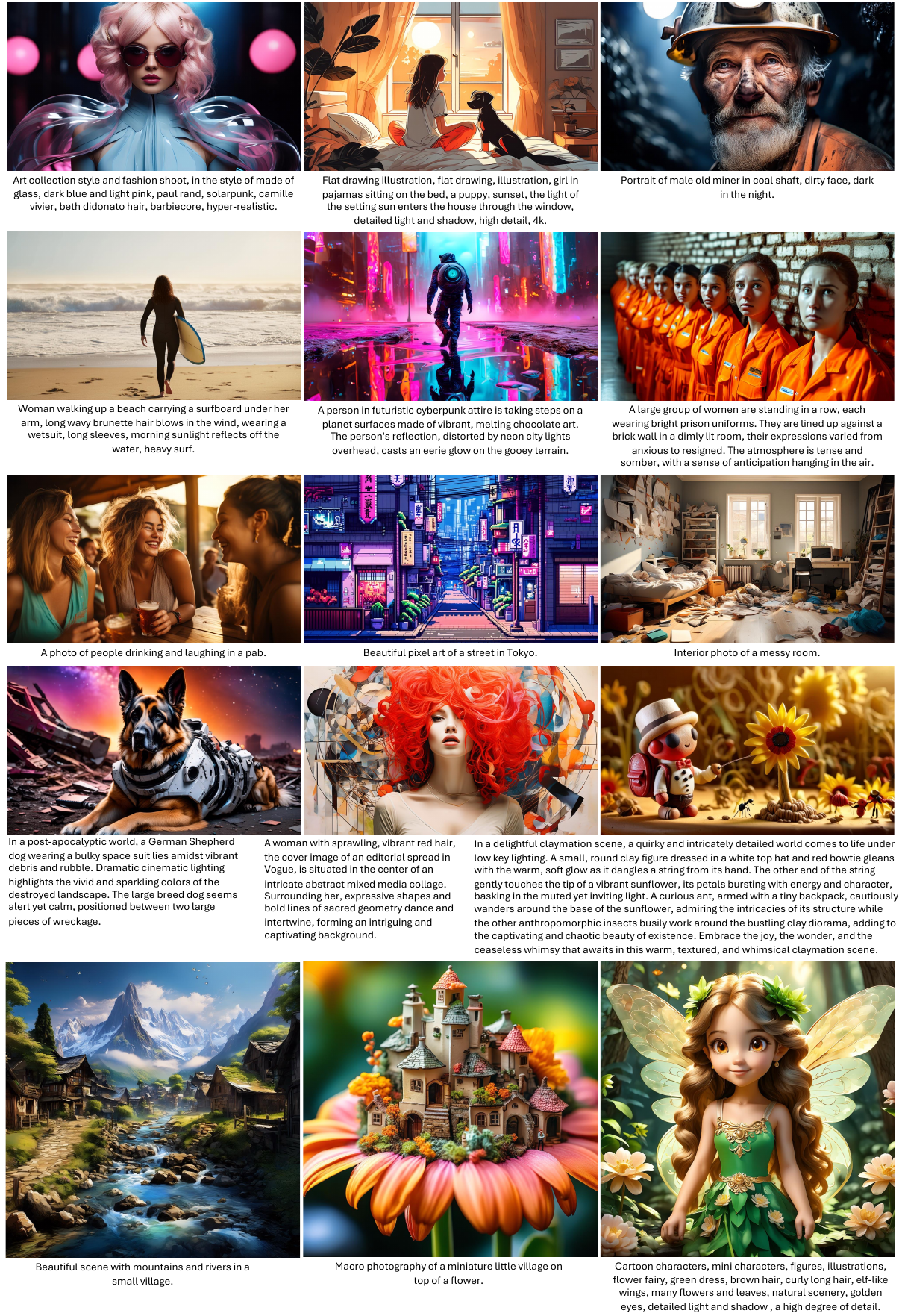}
  \vspace{-5pt}
  \caption{Text-to-1K image generation results produced by our Hydra-Transformer Hybrid model. The images are in resolutions 768$\times$1344 and 1344$\times$1344. Please zoom in for clearer visualization.}
  \label{fig:visualize_96x168_168x168}
\end{figure*}

\begin{figure*}[p]
  \centering
  \includegraphics[width=0.86\textwidth]{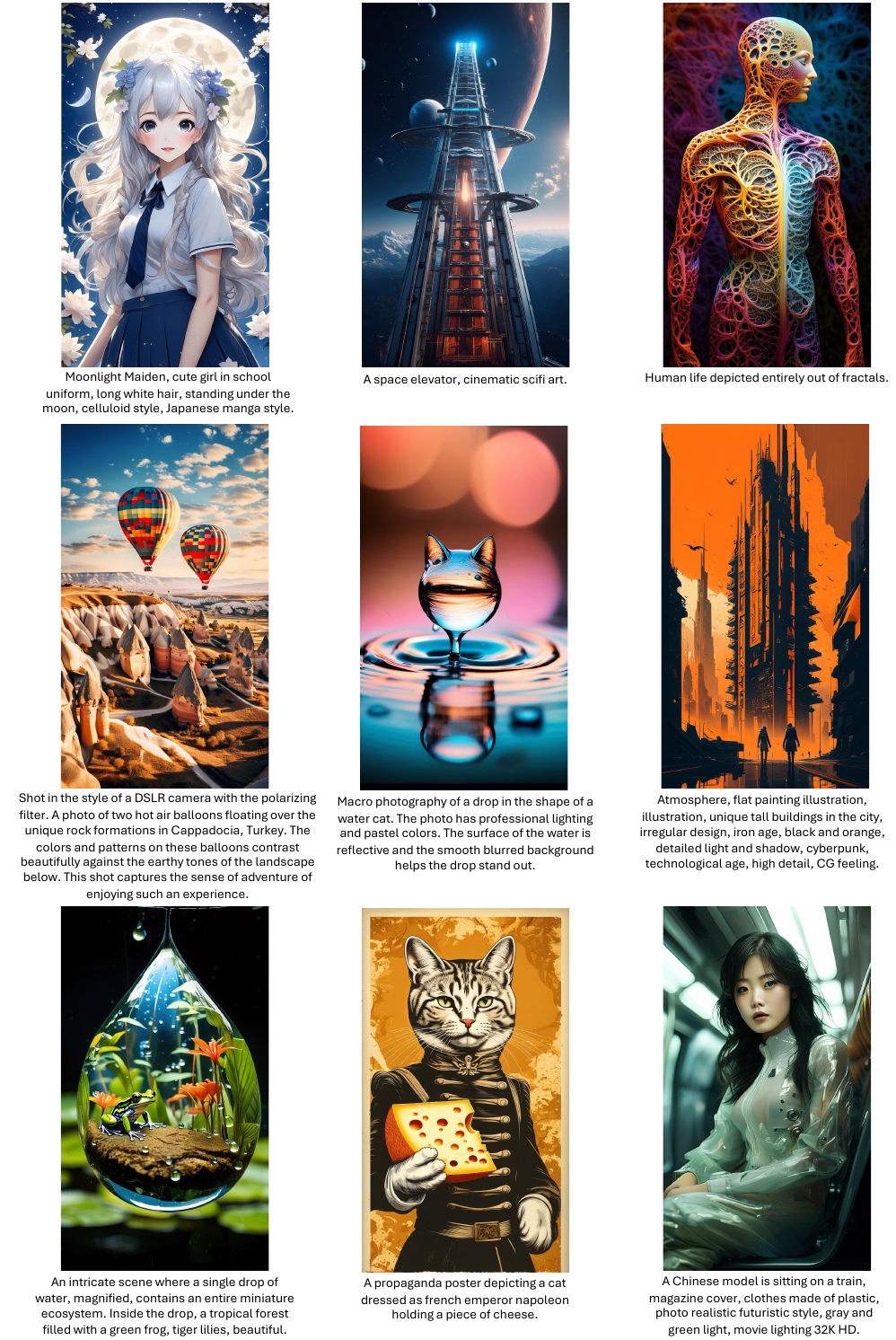}
  \vspace{-5pt}
  \caption{Text-to-1K image generation results produced by our Hydra-Transformer Hybrid model. The images are in resolutions 1344$\times$768. Please zoom in for clearer visualization.}
  \label{fig:visualize_168x96}
\end{figure*}

\begin{figure*}[p]
  \centering
  \includegraphics[width=0.89\textwidth]{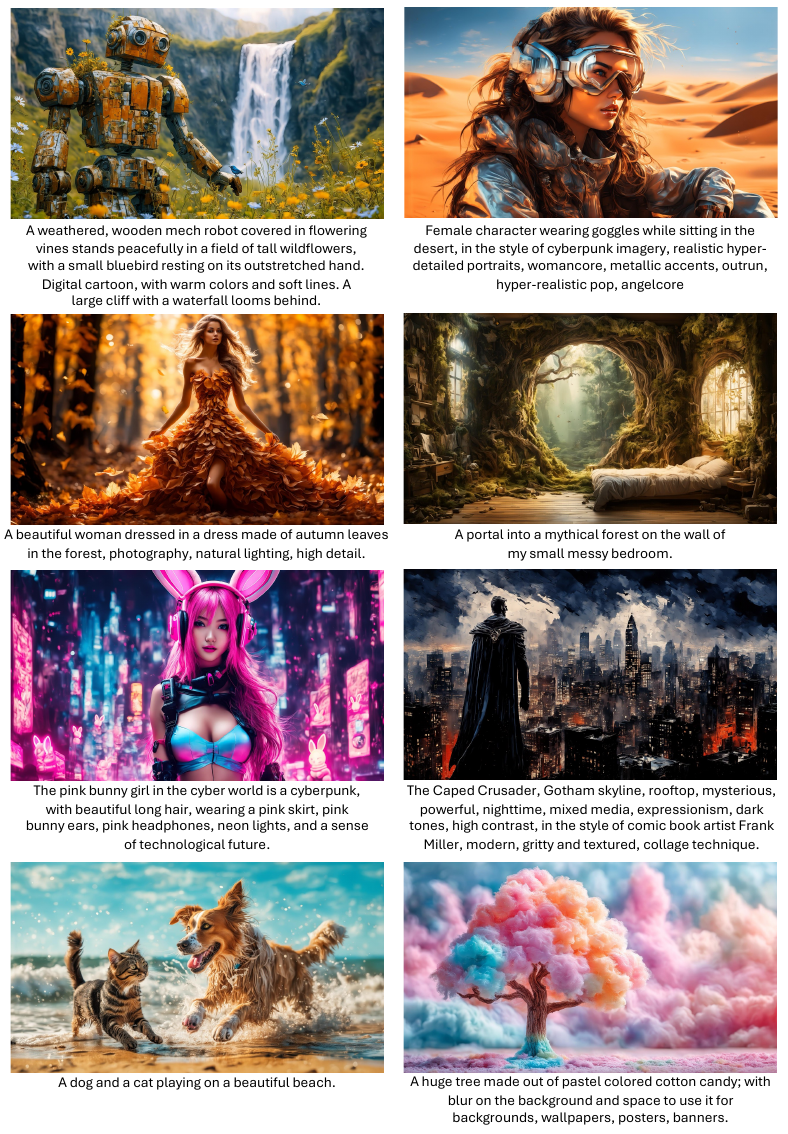}
  \vspace{-5pt}
  \caption{Text-to-2K image generation results produced by our Hydra-Transformer Hybrid model. The images are in resolutions 1440$\times$2560. Please zoom in for clearer visualization.}
  \label{fig:visualize_180x320}
\end{figure*}

\begin{figure*}[tp!]
  \centering
  \includegraphics[width=0.88\textwidth]{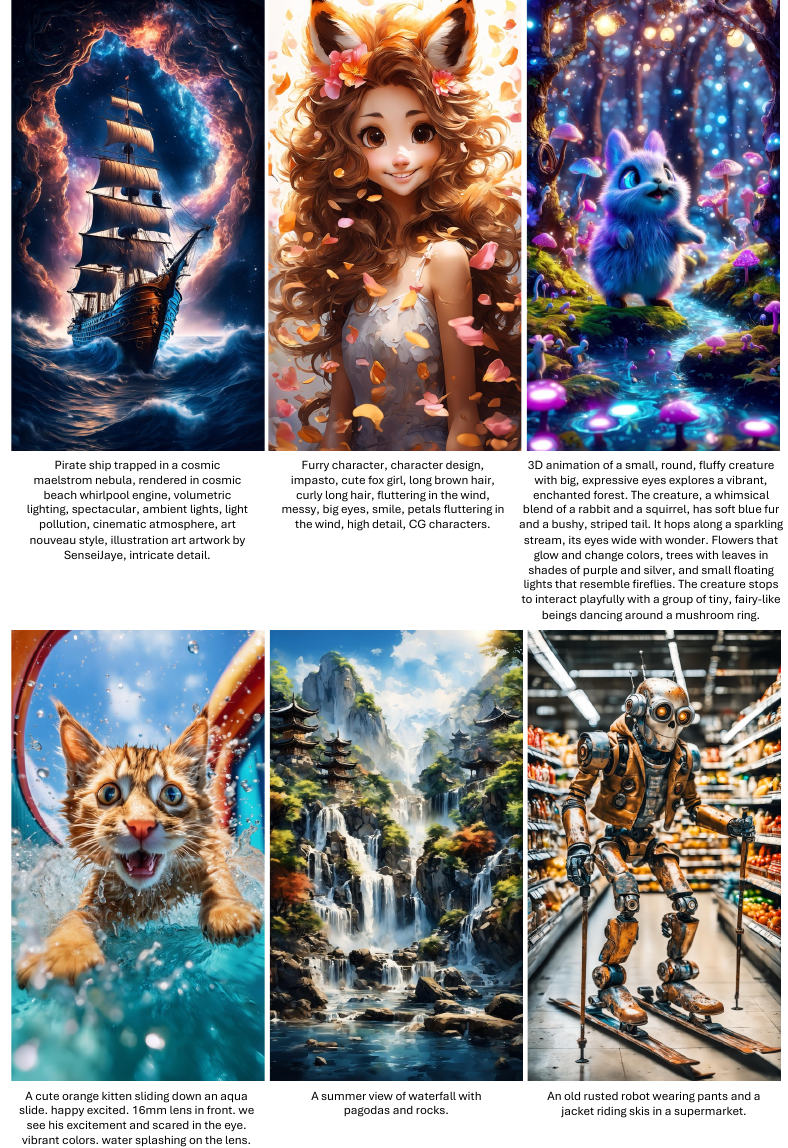}
  \vspace{-5pt}
  \caption{Text-to-2K image generation results produced by our Hydra-Transformer Hybrid model. The images are in resolutions 2560$\times$1440. Please zoom in for clearer visualization.}
  \label{fig:visualize_320x180}
\end{figure*}

\begin{figure*}[tp!]
  \centering
  \includegraphics[width=0.88\textwidth]{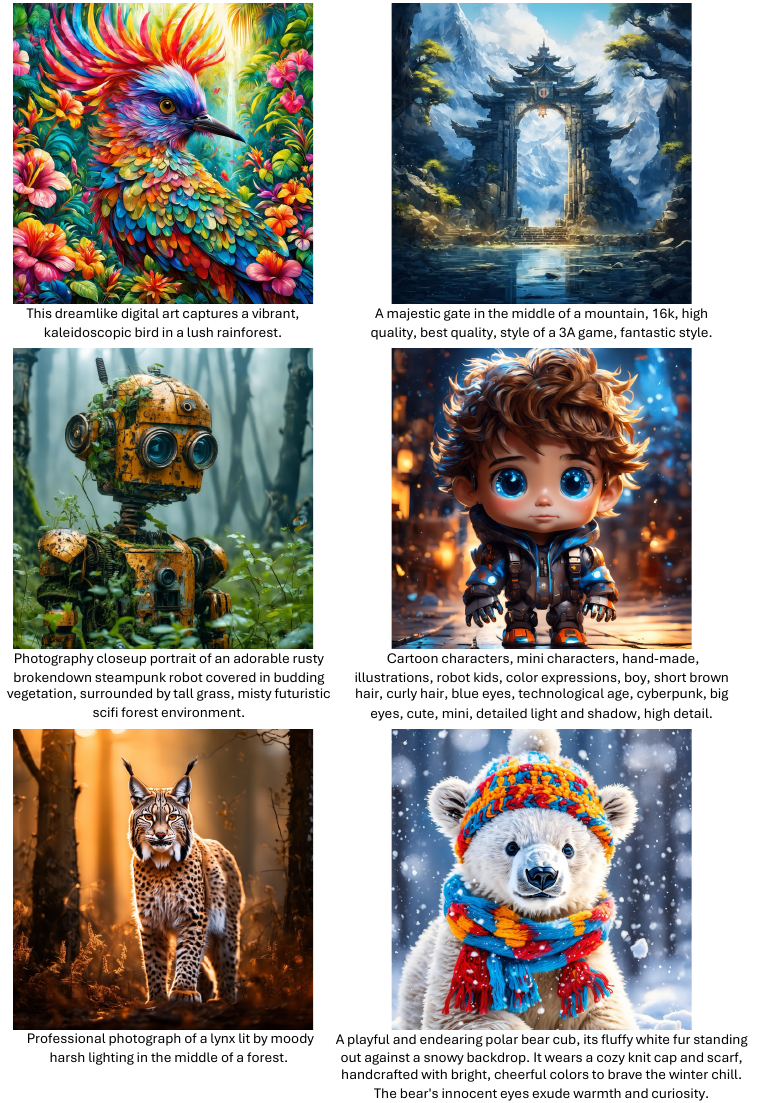}
  \vspace{-5pt}
  \caption{Text-to-2K+ image generation results produced by our Hydra-Transformer Hybrid model. The images are in resolutions 2560$\times$2560. Please zoom in for clearer visualization.}
  \label{fig:visualize_240x240}
\end{figure*}

\section{Prompts of Figure~\ref{fig:visualize_1}}

We provide the prompts for generating the images presented in Figure~\ref{fig:visualize_1} here. In a raster scan order: 

\begin{enumerate}
    \item A dog that has been meditating all the time.
    \item In a surreal office setting during a sunlit day, a man in a bizarre jelly mask, reminiscent of an oyster shell, occupies a full shot. He is seated at an office desk, his melancholic expression in contrast to the elaborate facades of yttrium yellow and zircon blue surrounding him.
    \item Full-body photograph of a beautiful female cyborg, suspended in a worn-out, cassette-futurism science lab. The scene exudes a gritty retro sci-fi aesthetic from the 70s and 80s, with outdated technology and weathered machinery. Dust particles float in the light, enhancing the vintage, worn-down atmosphere. The cyborg’s synthetic skin is torn from her down to her, revealing a complex but aged, metallic structure beneath. Her damaged breasts and partially exposed torso show signs of wear, with cracks and rust adding to the dystopian feel. Her body hangs lifeless from flickering, illuminated cords connected to her back.
    \item A bouquet of roses made of pastel ice crystals.
    \item A cat drinking a beer.
    \item Paper artwork, layered paper, colorful Chinese dragon surrounded by clouds.
    \item Beautiful neon lights forming the words ``Hydra'', glowing vibrantly.
    \item Filmic photo of a group of three women on a street downtown, they are holding their hands up the camera.
    \item A pink bunny girl in the cyber world, with beautiful long hair, wearing a pink skirt, pink bunny ears, pink headphones. Neon lights, and a sense of technological future.
    \item Half human, half robot, repaired human, human flesh warrior, mech display, man in mech, cyberpunk.
    \item Elephant amigurumi walking in savanna, a professional, blurry background.
    \item Polychrome particles and powders surrounding a whimsical child figure silhouette emerging out of a portal.
    \item A very cute little Shamrock bird in a mossy forest.
\end{enumerate}

\end{document}